# Causal AI-based Root Cause Identification: Research to Practice at Scale

A technical and formal exploration

Saurabh Jha, IBM Research, saurabh.jha@ibm.com
Ameet Rahane, IBM Instana, ameetrahane@ibm.com
Laura Shwartz, lshwart@us.ibm.com
Marc Palaci-Olgun, IBM Instana, marc.palaci-olgun@ibm.com
Frank Bagehorn, IBM Research, fba@zurich.ibm.com
Jesus Rios, IBM Research, jriosal@us.ibm.com
Dan Stingaciu, IBM Instana, Dan.Stingaciu@ibm.com
Ragu Kattinakere, IBM Instana, raguk@ca.ibm.com
Debasish Banerjee[1], Guild Systems Inc., debasish.banerjee@guildsys.com

# Introduction

Modern applications are increasingly built as vast, intricate, distributed systems. These systems comprise various software modules, often developed by different teams using different programming languages and deployed across hundreds to thousands of machines, sometimes spanning multiple data centers. Given their scale and complexity, these applications are often designed to tolerate failures and performance issues through inbuilt failure recovery techniques (e.g., hardware or software redundancy) or external methods (e.g., health check-based restarts). Computer systems experience frequent failures despite every effort: performance degradations and violations of reliability and Key Performance Indicators (KPIs) are inevitable. These failures, depending on their nature, can lead to catastrophic incidents impacting critical systems and customers. Swift and accurate root cause identification is thus essential to avert significant incidents impacting both service quality and end users.

In this complex landscape, observability platforms that provide deep insights into system behavior and help identify performance bottlenecks are not just helpful—they are essential for maintaining reliability, ensuring optimal performance, and quickly resolving issues in production. The ability to reason about these systems in real-time is critical to ensuring the scalability and stability of modern services.

To aid in these investigations, observability platforms that collect various telemetry data to inform about application behavior and its underlying infrastructure are getting popular. Examples of such platforms include IBM Instana, Dynatrace, Datadog, AppDynamics, New Relic, Grafana

---

[1] Formerly at IBM

Labs, and several others. In addition to providing rich telemetry data, including traces, metrics, and logs, most of the available Application Performance Monitoring (APM) tools purport to determine the root causes of system issues and pinpoint them to the Site Reliability Engineers (SREs).

IBM Instana uniquely stands out compared to several other APM tools in using 'causal AI' to surface the root causes of the system problems to the SREs in near real-time. Instana's Root Cause Identification (RCI) algorithm honors the adage: 'correlation is not causation' by focusing on 'causation' and not 'correlation.'

The paper will delve into Instana's failure diagnosis capabilities, with special emphasis on its implementation of the RCI algorithm and its theoretical aspects at a high level of abstraction. We will highlight the usefulness of Instana's RCI implementation, citing real-life examples.

For an overall description of IBM Instana, the reader can consult the latest documentation at https://www.ibm.com/docs/en/instana-observability/current.

# Motivating examples

Distributed tracing is widely believed to be sufficient to identify the root cause. While tracing is invaluable, in the following examples, we illustrate why distributed tracing by itself may be inadequate for determining root causes in complex production environments.

## Example 1: Misattribution from connection pool exhaustion

In the scenario depicted in Figure 1, a user issues an HTTP request to an application server (denoted **appserver**), which is a WebSphere-based service. The **appserver** then attempts to create a new database connection to **mydb2**, a DB2 instance. However, the trace reveals a failure indicating that **mydb2** could not allocate a connection. A cursory interpretation of this trace might lead Site Reliability Engineers (SREs) to conclude that mydb2 is at fault and ought to be scaled or reconfigured to accommodate additional connections. The logs of **mydb2** clearly indicated that the connections were exhausted.

Upon closer inspection, it turns out that **mydb2** was correctly configured for 70 active connections, but those connections were exhausted due to a deadlock condition in **appserver**. This deadlock triggered an uncontrolled surge of connection requests, ultimately saturating the DB2 connection pool. Consequently, the real issue lay within the application server rather than the database tier. In fact, other dependent services that also required database access began to fail in a cascading manner. Diagnosing such deadlocks in a time-constrained, high-pressure production setting often requires deep insight and experience into both the application and the underlying infrastructure metrics (e.g., CPU and memory usage, thread states). This example also highlights a common misconception: the last node (the 'leaf') in a failing trace is the true root cause.

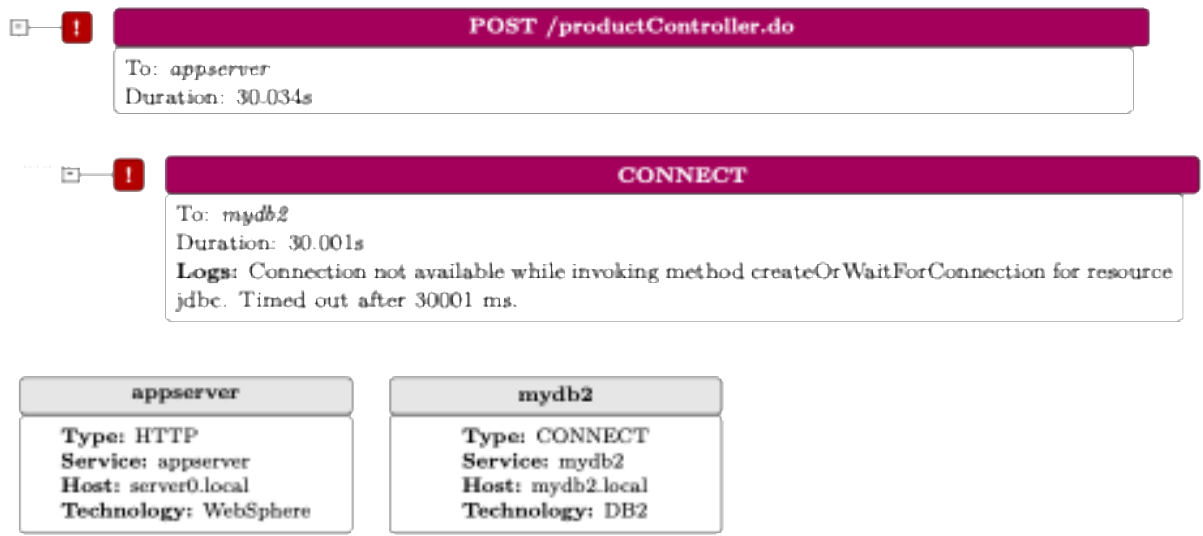

Figure 1. Trace of a failed request from appserver to mydb2.

## Example 2: Ambiguity in service-level vs. replica-level failures

Figure 2 depicts a failing trace in which a **frontend** service interacts with the **catalogue** service via the endpoint **/api/product/sku1**. The trace concludes with an HTTP 503 "service unavailable" error. An immediate inference might be that the **catalogue** service is failing outright. However, in this environment, **catalogue** has three replicas in production. The question arises: is the entire service (i.e., its code or configuration) faulty, or is only one replica—identified here as `catalogue-abceg-2434`—exhibiting abnormally high resource consumption or other anomalies?

To resolve this, an SRE must look at additional traces or system metrics across all catalogue replicas. For instance, Figure 3 shows a successful trace involving the same frontend and the **catalogue** service but routed to a different replica (`catalogue-ddeiew-18434`), thereby illustrating that not all replicas are impacted. In large-scale distributed systems, numerous dimensions may have to be considered—such as different pods, containers, virtual machines, APIs, operating systems, or deployment configurations. SREs may face an inherently high-dimensional root-cause analysis problem, which is impractical to tackle manually on each occurrence of failures.

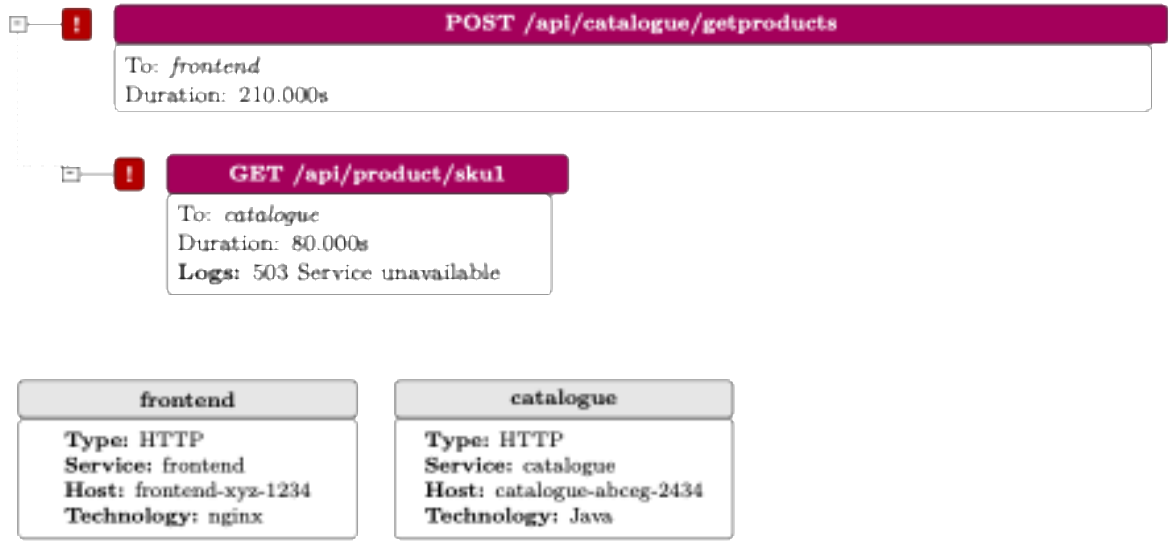

Figure 2. Trace of a failed request from frontend to the catalogue service

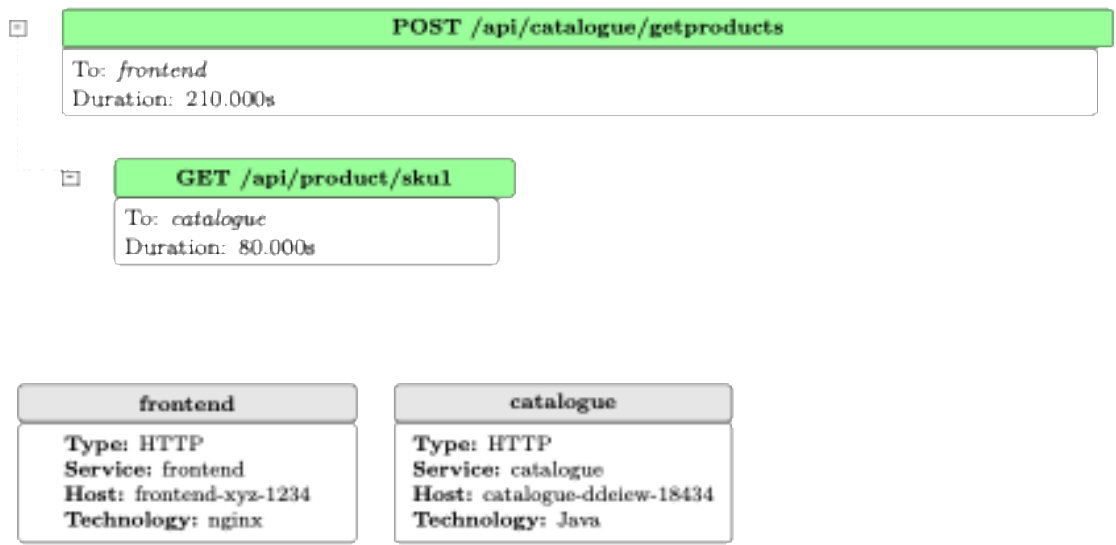

Figure 3. Trace of a successful request from frontend to the catalogue service

These two examples underscore that naive reliance on distributed tracing alone can lead to spurious or incomplete conclusions about the root cause of failures. Tools incorporating additional causal inference—such as IBM Instana's Root Cause Identification (RCI) engine—help SREs quickly distinguish between true root causes and symptomatic side effects in large-scale distributed environments. Subsequent sections of this paper discuss the principles underlying Instana's RCI algorithm and illustrate how it mitigates the inherent diagnostic challenges in complex distributed environments highlighted by these scenarios.

# Background

Academics and practitioners have extensively studied failure diagnosis in computers and related systems. As a result of their effort and contributions, we now have powerful frameworks and theoretical machinery to continue the work on accurate failure diagnosis and making computer systems more robust through proper remedial actions.

The basic concepts of dependability and taxonomy of fault-tolerant computing are laid out nicely by Avizienis et. al. at https://ieeexplore.ieee.org/document/1335465. Readers can consult the paper for a thorough introduction to the area.

IBM Instana's causal AI implementation is based on Pearl's groundbreaking work on causality. For an excellent overview of causality, readers can consult the Book of Why and Causal Inference in Statistics: A Primer. The present paper will discuss Instana's causal AI implementation using several well-known concepts and symbolisms of statistics and probability. If necessary, readers can consult any standard available material on statistics and probability; Probabilistic Graphical Models is an excellent choice.

## Definitions

We introduce several key definitions to help readers understand the basic concepts and taxonomy of dependable computing before delving into the main topic: "How does IBM Instana help resolve real-world incidents?"

**System**: A *system* is a set of connected components that interact to deliver one or more services. A component by itself can also be viewed as a system – such a component is considered atomic. Identifying atomic components may depend on the specific circumstances and the investigating Site Reliability Engineers (SREs). For example, in a cloud-native application, individual microservices can be considered atomic components, or depending on the need of the situation, an SRE can drill down to investigate individual pods or containers or even lower code-level modules in search of atomic components appropriate for trouble shootings. As a general suggestion, SREs/diagnostic frameworks should define components as atomic as they see fit for specific incident resolutions and recoveries.

**User**: A user is an individual or a system that has ability to consume actions and tasks provided by the system.

**Service**: the service provided by a system are tasks and actions performed for its user(s) based on a mutual agreement, called Terms of delivery.
Service delivery occurs at the service interface, which is part of the provider's system boundary.

**Failure**: A service *failure* occurs when a user cannot correctly use the service as expected or as specified in the term of the service. The period during which the service does not comply with terms of delivery is called the service outage. The transition from non-compliant service to compliant service is called a service restoration. The deviation from the correct service may take different forms, called service failure modes, which are ranked according to failure severity.

**Error**: At the time-of-service failure one or more external states of the service deviate from the correct service state. This deviation in service state is called an *error*.

**Fault**: A *fault* is the adjudged or hypothesized cause of an error. Faults can
be internal or external to a system. Academicians and practitioners have defined several fault categories. A fault can simultaneously belong to multiple categories. This paper focuses on 'operational faults' that occur during the operational life cycles of systems. A component can simultaneously have one or more fault, error, and failure states.

The above three definitions should make it clear that *errors* occur because of *faults*, which result in service *failures*. In this paper, we may use the terms fault, error, and failure interchangeably when there is no chance of confusion.

**Fault Localization**: Fault Localization is the process of identifying the fault that caused a service failure for a given system and its atomic components.

In real life, faults can be transient or permanent. A transient fault happens occasionally, resulting in a transient failure of services. Unexpected input to a system, temporary scarcity of computing resources, and sudden spikes in load are the leading causes that can contribute to transient failures of services. Sporadic transient failures may not demand immediate attention from SREs to prevent their occurrence, though SRE may choose to consider preventing their occurrence. Permanent failures, in general, have to be addressed by SREs urgently to restore the services. For simplicity, in this paper, we assume that restarts of the relevant atomic components will restore services. However, it should be noted that in real life, restoring services may need the application of emergency code fixes, configuration changes, and other specific measures before restarting the relevant atomic components.

# Failure propagation

In a network of interconnected services, a single fault can cascade across multiple components, ultimately manifesting as a system-wide failure from the user's perspective. Figure 4 illustrates this phenomenon using System E, a simplified shopping platform. When a fault arises during a transaction, users perceive the entire system as failing. However, the root cause can be pinpointed in the Balance Check Container (System C) from the perspective of an SRE with full observability into System E. This localized fault propagates through the system, triggering service disruptions in Balance Check, the web server (System B), the payment gateway, and the web server (System A), thereby propagating the failure throughout System E.

By contrast, when SREs have only partial visibility—for example if they can observe only System A—they might attribute the anomalies to the payment gateway in System A without recognizing the upstream fault in System C. Modern observability platforms may still allow SREs to intercept outgoing calls from System A and trace problems to System D. However, they cannot directly see the underlying issue in System C. Finally, if one's visibility is constrained solely to System C, the event appears as a localized fault, error, and failure within that container, providing no insight into its effects on other subsystems.

Hence, fault localization depends on the boundaries of visibility and on how responsibilities are partitioned. A component can be considered 'atomic' or fully decomposed depending on the observer's vantage point: for instance, SREs responsible for System A may view System D as atomic, whereas SREs managing System E can fully observe System D and thus do not treat it as atomic.

This example underscores how the scope of observability governs failure diagnosis. The broader the visibility, the more precisely a root cause can be identified; conversely, limited visibility compresses the fault, error, and failure into a single subsystem, treating it as an atomic black box.

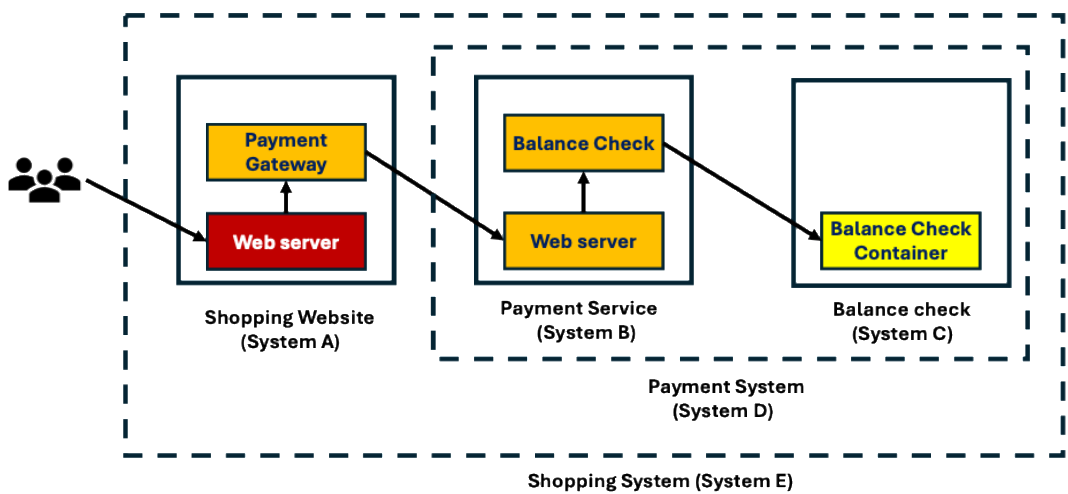

Figure 4. Failure propagation in a shopping system

# Failure diagnosis and remediation

To remediate system failures, we need to accurately localize the faults. Once a fault is identified, SREs can take remediation action to restore the affected service. This section elaborates on the fundamental task of fault localization and consequent remediation actions.

## Fault localization and root cause

One of the most crucial steps in operational failure resolution is identifying the root cause—or fault—underlying a system failure. However, this notion of a root cause is only meaningful once the system's boundaries have been established. In other words, for a defined system, the root cause is the fault responsible for the observed service failure. Yet, simply finding this root cause may not suffice to restore the service because faults can propagate through multiple components. Moreover, such propagation can be transient or permanent. For instance, a brief outage in a database might permanently incapacitate a client application if the latter enters an unrecoverable error state.

For example, consider SREs managing the shopping system (System E) from Figure 4, where each colored box is treated as an atomic component. In the previously described fault scenario, they would first attempt to restart the Balance Check Container (System C). If this restart fails to resolve the issue, they might need to restart System B's Balance Check component. Meanwhile, other atomic components would self-correct if the fault's impact on them were transient, sparing further remedial action. Hence, only two atomic components may require actual intervention in this situation.

An SRE may also treat multiple subsystems as a single atomic entity for remediation. Suppose SREs consider System E in Figure 5 to be composed of two atomic components: System A and System D, where Systems B and C of Figure 4 are conceptually merged into System D. If a restart operation is available at the level of System D, the SREs might restart the entire payment subsystem as a single unit. By considering System D as atomic, only one remediation step is required—demonstrating how system boundaries and the chosen level of observability can radically alter the scope of fault resolution.

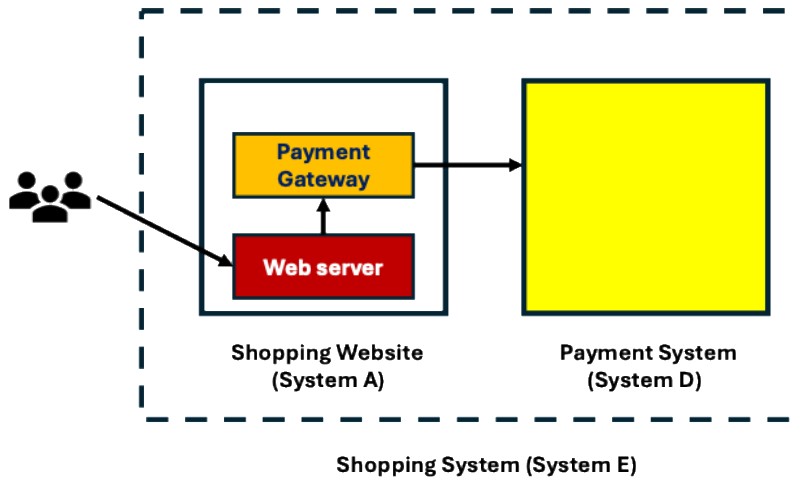

**Figure 5.** Failure propagation in a shopping system — with redefined system boundaries

The fault localization in a system depends on how the 'system' and its 'atomic components' are defined. Unfortunately, there is no universally agreed taxonomy of the system and its atomic components in the published academic literature or the documents of monitoring and observability platforms in the market. SREs define the system boundary based on their responsibilities and visibility into the system. For a 'fault,' this often results in ambiguities or impreciseness in the "identified root cause" by any method, manual or automatic.

## Practical insufficiency of the fault localization

Failure diagnosis may appear to be a straightforward extension of fault localization—essentially, iteratively identifying the faulty component by delving deeper into each atomic element and treating it as a system in its own right. However, limited visibility into these individual components often poses significant challenges in practice. In real-world enterprise environments, not every part of an application is monitored. Cost, performance overhead, lack of suitable tools, or other constraints frequently hinder comprehensive monitoring. For instance, Gartner predicted in 2021 that by 2023, Application Performance Monitoring (APM) solutions—providing tracing capabilities—will be enabled on only about 20% of enterprise applications.

To address this shortage of observability data, observability vendors and customers commonly leverage pattern recognition and anomaly detection techniques. These methods often rely on carefully curated training datasets, log and metric anomaly identification, or handcrafted labeled datasets. Unfortunately, obtaining sufficient high-quality labeled data to build robust machine learning models is difficult because actual component failures are relatively infrequent in production environments. Anomaly detection alone can be ambiguous: an anomaly could be a positive signal during promotional periods in an eCommerce application but a negative one under normal conditions. Relying solely on anomalies can, therefore, yield false positives.

In real-world scenarios, we should combine multiple approaches to reduce our dependence on models that require extensive training or that tend to generate false positives. Instana's fault diagnosis, rooted in Causal AI, is augmented by machine learning or statistical anomaly detection methods in areas lacking observability. The causal AI–based fault localization model from Instana does not require any training and is broadly applicable. It also reports when its confidence is low due to insufficient visibility; at that point, site reliability engineers (SREs) can gather additional data—using complementary techniques like anomaly detection, pattern matching, or previous fault localization successes—to pinpoint the issue.

Instana's rigorous theoretical foundations help bridge the gap between ideal scenarios with rich observability and real-life systems with limited observability. In internal experiments involving production-scale enterprise applications, Instana successfully localized faults by accurately identifying root causes in approximately 90% of the cases.

The next section explores system visibility's importance in greater detail.

# Causal AI-based fault localization – a novel approach

Modern distributed systems consist of numerous interconnected services deployed across diverse hosts or cloud instances. When a request fails, ferreting out its root cause can be daunting: multiple components might fail simultaneously, or hidden resource pressures could degrade performance. Simple correlation-based methods (e.g., "Requests to Service A slow down whenever CPU usage on Host $N_1$ spikes") risk misdiagnosis of the true root cause. The **Pearlean framework** of **Causal Inference** techniques offers a theoretically sound mechanism to identify *why* a failure occurs. **Structural Causal Models (SCMs)** let us formally represent dependencies and confounders, while **Judea Pearl's *do*-calculus** helps us reason about the effect of "*forcibly* failing" or "intervening on" a component. However, building a complete SCM for a large-scale system often requires extensive data collection and sophisticated modeling of partial dependencies or continuous resource usage.

This section presents a simplified causal method grounded in **Beta-Binomial** inference. We assume fail-stop faults (i.e., each component is either 'healthy' or 'failed') and use an "AND" dependency rule (a request fails if *at least one* of its required components fails). These assumptions let us derive a *computationally efficient* **Bayesian model** for fault localization. We also discuss the limitations of this approach and propose ways to handle quorum-based dependencies, continuous metrics, and time-evolving states in future work.

## A simple example

As discussed earlier, faults propagate through the system. They can be transient or permanent depending on their impact on each atomic component. Hence, we adopt the process of iteratively identifying faults, assuming that the SREs will fix one fault at a time.

To help the reader understand the present work's core novelty, we first define an example distributed system and list a few simplifying assumptions about it. We then examine how faults can occur and propagate within the system.

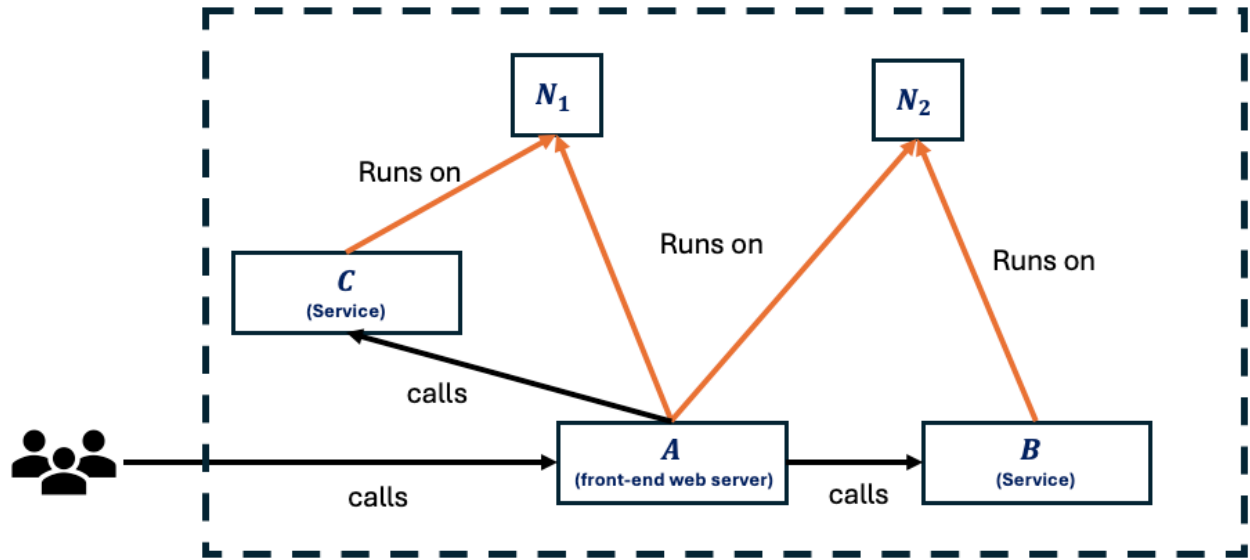

**Figure 6. Example System – a sample distributed system**

Our Example System, shown in Figure 6, consists of Service $A$, a front-end web server with two replicas—one replica running on the host $N_1$ and another on the host $N_2$; Service $B$, an upstream service running solely on the host $N_2$; and Service $C$, another upstream service running exclusively on the host $N_1$. The services are hosted on physical or virtual machines, represented by hosts $N_1$, and $N_2$. We will use the Example System as a running example throughout this paper.

> ### Assumptions A1
> A1.1    We assume the services are running directly on Kubernetes hosts. Adding containers, pods, virtual machines, etc., will not add additional insights for understanding. Our general approach remains valid in such cases
> A1.2    Placement of services to the hosts is static, i.e., services do not migrate automatically to others hosts.
> A1.3    There is no automatic restart feature. If a component fails, it must be restarted manually.

The frontend web server $A$ supports two types of requests. Requests
- which require interaction with Service $B$ (path: A -> B), and
- which require interaction with Service $C$ (path: A -> C)

Given the possible paths in the system, we can enumerate the request types as follows:

- $R_1$: Replica of $A$ running on $N_1$ calls $B$ running on $N_2$.
- $R_2$: Replica of $A$ running on $N_1$ calls $C$ running on $N_1$.
- $R_3$: Replica of $A$ running on $N_2$ calls $C$ running on $N_1$.
- $R_4$: Replica of $A$ running on $N_2$ calls $B$ running on $N_2$. To further simplify our discussion, let us assume that request type $R_4$ is not allowed in our example system because of a circuit breaker.

Figure 7 depicts the same Example System of Figure 6, explicitly showing the flow of the requests mentioned above.

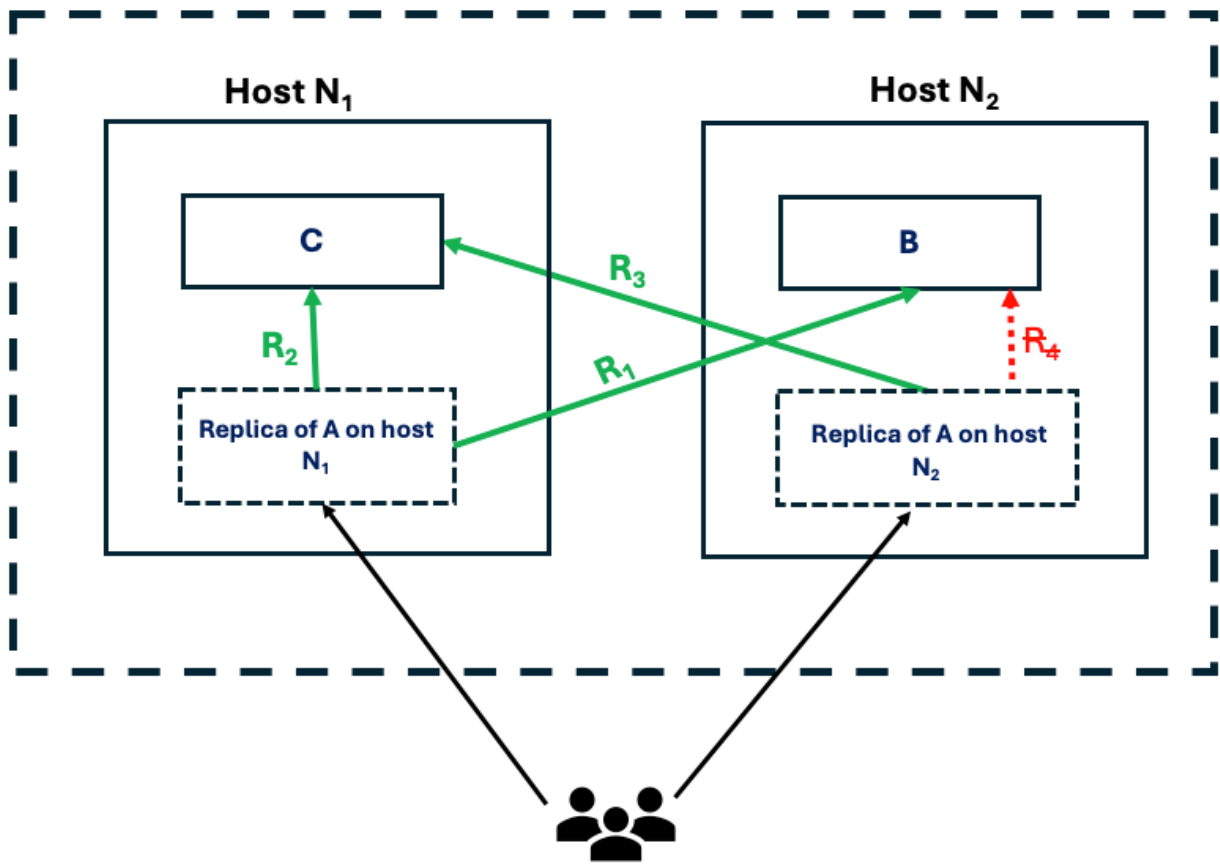

Figure 7. The flow of requests in the Example System

Figure 6 shows many different relations (e.g., 'Runs On' and 'Calls') that can be formally modeled using mathematical structures like directed graphs, potentially including containment relationships. However, precisely defining these relationships often involves significant ambiguity. For simplicity and mathematical clarity, we focus solely on modeling the "depends on" relationship, effectively projecting all other relationships onto this single relationship. For example, the successful completion of a request of type $R_1$ depends on the elements of the set $\{A, B, N_1, N_2\}$. This is because the request requires the invocation of a replica of the web server $A$ running on the host $N_1$ to service $B$, which is itself hosted on the host $N_2$. We may refer to the set as the 'dependency set' of the request $R_1$. In this paper, we often denote the dependency set corresponding to a request type $R_j$ in a system as *DependsOn(R_j)*.

Execution of a request will be associated with symptoms or observable errors. Note that the absence of any observable errors in the execution of a request can also be considered a (trivial) symptom that indicates the success of the execution.

For a request type $R$, $Y_R$ denotes the success or failure of the request $R$. $Y_R$ can have two possible values: 0 and 1 --- 0 indicates the failure, while 1 indicates the success of executing a request of type $R$.

## Fault scenarios

Let's examine three different fault scenarios to understand how faults can propagate through this system and impact requests.

### Scenario 1: Fault in service $B$

1. A fail-stop fault occurs in the Service $B$ running on the host $N_2$.
2. The fault impacts only request type $R_1$, as Service $B$ belongs to the dependency set of $R_1$.
3. Service $A$ continues to operate. Thus, $R_2$ and $R_3$ are not impacted.
4. Fixing service $B$ recovers the system entirely.

Timeline:
- $t_0$: System functioning normally.
- $t_1$: Fault occurs in Service $B$.
- $t_2$: $R_1$ requests fail, $R_2$ and $R_3$ requests continue to work normally.
- $t_3$: Fault in Service B is fixed.
- $t_4$: Requests of all three types: $R_1$, $R_2$ and $R_3$ succeed.

This scenario demonstrates how a fault in an upstream service can impact specific request types ($R_1$ in this case) without causing a complete failure across all request types. The different failure patterns across various request types illustrate the important principle of 'differential observability,' discussed later in this section.

### Scenario 2: Fault in service $A$

1. A fail-stop fault occurs in Service $A$, affecting its replicas on $N_1$ and $N_2$.
2. This fault impacts all request types: $R_1$, $R_2$ and $R_3$.
3. Fixing Service $A$ recovers the entire system.

Timeline:
- $t_0$: System functioning normally.

- $t_1$: Fault occurs in Service $A$.
- $t_2$: Requests of all three types: $R_1$, $R_2$ and $R_3$ fail.
- $t_3$: The fault in Service $A$ is fixed.
- $t_4$: Requests of all three types succeed, and the system functions again normally.

This scenario illustrates how a fault in a front-end service can have a widespread impact on user experience without involving other back-end services, even though no fault propagation to Services B or C occurs here.

### Scenario 3: Fault in host $N_1$

1. A fail-stop fault occurs in host $N_1$.
2. This causes fail-stop failures in the Service $A's$ replica on $N_1$ and Service $C$ on $N_1$.
3. This fault impacts all request types: $R_1$, $R_2$ and $R_3$.
    - $R_1$: fails since $A$ is unavailable, because $N_1$ is unavailable.
    - $R_2$: fails since $A$ and $C$ are unavailable, because $N_1$ is unavailable.
    - $R_3$: fails since $C$ is unavailable, because $N_1$ is unavailable.
4. Fixing $N_1$ doesn't automatically fix the replica of Service $A$ running on $N_1$ and Service $C$ running on $N_1$; they must be restarted separately. Recall there is no automatic restart feature in our Example System.

Timeline and iterative fix process:
- $t_0$: System functioning normally.
- $t_1$: Fault occurs in host $N_1$.
- $t_2$: Requests of all three types: $R_1$, $R_2$ and $R_3$ fail.
- $t_3$: Host $N_1$ is brought to a healthy state, but the replica of $A$ and $C$ in $N_1$ remain in the 'fail-stop' states.
- $t_4$: Requests of all three types: $R_1$, $R_2$ and $R_3$ are still failing.
- $t_5$: Service $C$ is restarted
- $t_6$: $R_3$ requests succeed; requests of types $R_1$ and $R_2$ still fail since the replica of $A$ in $N_1$ has not been fixed.
- $t_7$: The replica of Service $A$ on $N_1$ is restarted.
- $t_8$: Requests of all three types succeed, and the system functions normally.

This scenario illustrates the complexity of fault propagation in distributed systems; a single point of failure can have cascading effects that require multiple steps to resolve fully.

Scenario 3 is interesting. Even after fixing the host $N_1$, none of the symptoms disappear. The replica of Service $A$ and Service $C$ running on host $N_1$ are essentially independent simultaneous faults that require independent mitigation. Moreover, due to failure propagation paths and our system topology, two independent faults, $A$ running on $N_1$ and $C$, are no different from one single fault on $N_1$. This scenario highlights several challenges that SREs may experience in the real world:

1. Indistinguishability of faults in terms of their symptoms (impacts).

2. The order of fixing. Which fix should be prioritized first in the remediation journey -- $N_1$, or the replica of $A$ on $N_1$, or $C$ on $N_1$, or all three simultaneously?

Expert SREs will recognize that the replica of $A$ on $N_1$ and C are both running on $N_1$; hence, they should prioritize fixing the actual fault, which is $N_1$. Once that is fixed, they will recognize, that $C$, the leaf node in the example system's topology, should be fixed first and, finally, $A$ should be fixed. Our methodology models the dependencies through a causal reasoning framework to help SREs quickly identify faults.

The order of fixing root causes matters for concurrent root causes, like the one here. Instana's theoretical framework does not guide SREs about the preferred order to fix concurrent root causes. Presently, we use a heuristic based on the probable impact of the root causes on the system to suggest an order to fix such root causes. We plan to improve the feature of ordering the mitigations of concurrent root causes in future releases.

## Important takeaways

These scenarios highlight several important considerations:
1. **Influence network**: Figure 8 pictorially captures in a set-theoretic fashion the influence of each component on the request type $R_j$, where $Y_{R_j} \in \{0, 1\}$. $Y_{R_j} = 0$ indicates request failure and $Y_{R_j} = 1$ indicates request's success.
2. **Differential impact**: Different faults produce distinct patterns of request failures, forming the basis for spatial differential observability and fault localization.
3. **Fault propagation paths**: Faults can propagate along specific paths in the system, causing cascading failures that must be addressed in a specific order.
4. **Partial failures**: Redundancy (e.g., multiple replicas of Service $A$) can mask faults, making them harder to detect and localize -- refer to the timeline entry $t_6$ in Scenario 3, where the call from the replica of $A$ in $N_2$ to $C$ in $N_1$ succeeds, though the replica of $A$ in $N_1$ is in fail-stop mode.
5. **Recovery process**: Fixing a fault doesn't always immediately restore system functionality, especially when the fault has propagated.

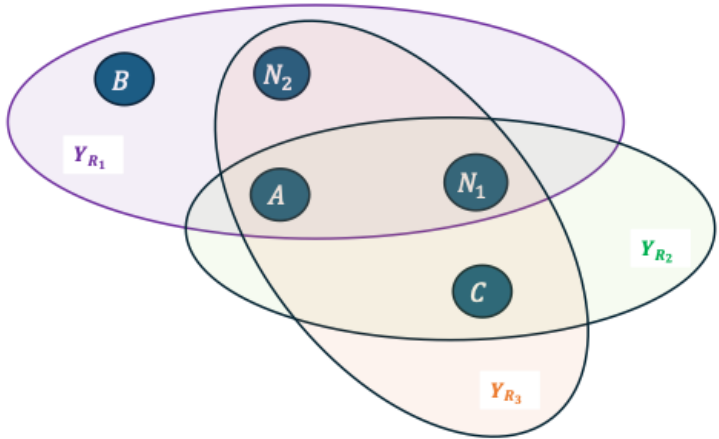

**Figure 8. The influence of each component on the success or failure of a request**

## Principle of Differential Observability

Faults can be localized by leveraging both **spatial** and **temporal differential observability**.

- **Spatial differential observability** exploits the premise that distinct faults within a system will manifest as unique failure patterns *across different request types*. Analyzing these variations in failure patterns across the "space" of request types helps pinpoint the location of faults. However, achieving this level of observability is not a given; it is influenced by the system's topological structure and the heterogeneity of request types.
- **Temporal differential observability** extends the above concept by considering the *time dimension*. It posits that different faults may also exhibit distinct temporal patterns in their failure behavior. For example, one fault might cause intermittent failures that occur with increasing frequency over time, while another might trigger a burst of failures followed by a period of apparent stability. By analyzing these time-varying characteristics of failures, additional information can be gleaned to aid in fault localization.

We use our Example System of Figure 6 to explain the essence of differential observability. To simplify the analysis, we'll focus exclusively on the 'fail-stop' failures and 'spatial differential observability.'

> **Assumptions A2**
> **A2.1** **Scope: Fail-Stop Failures Only:** This analysis is limited to fail-stop failures, in which components cease functioning entirely upon failure. Thus, the implications for temporal differential observability are minimized.
> **A2.2** **Scope: Single Fault Scenario:** The discussion is further limited to scenarios with a single active fault. These simplifications help readers understand the fundamental concepts without restricting the approach's applicability to a broader scope.

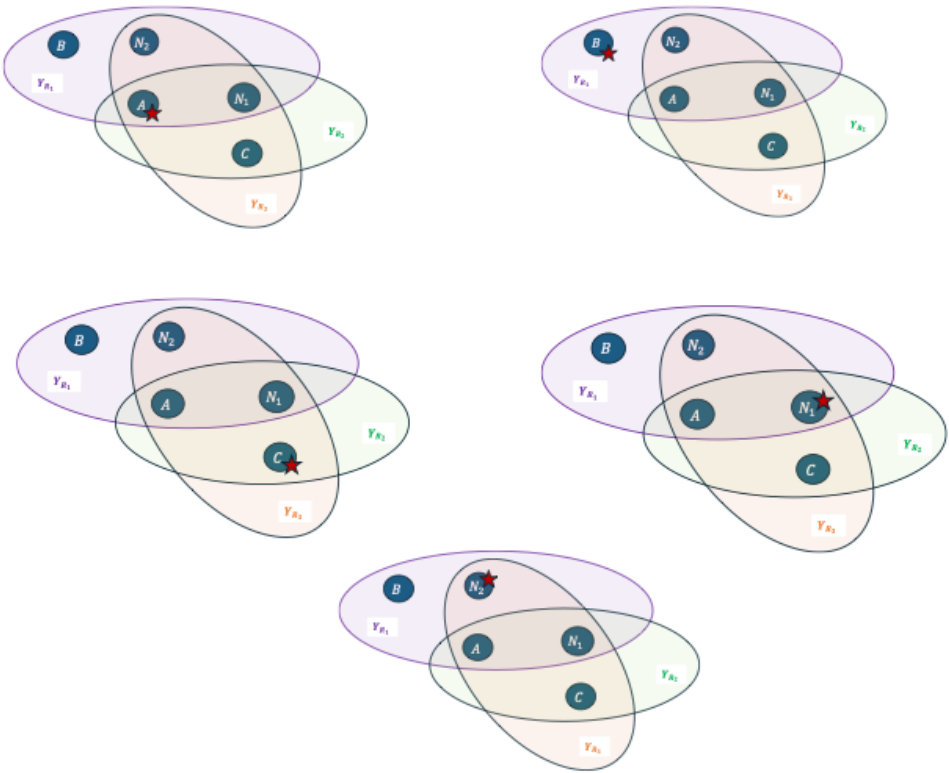

**Figure 9. Possible single component faults of the Example System**

Figure 9 shows all possible faults, marked with a red star, of the Example System of Figure 7 when faults are restricted to single atomic components. It is easy to see the impact of the faults on the valid request types by observing the **Venn diagrams** of the dependency sets of the request types. For example, when Service $A$ is at fault, all three request types will be affected since Service $A$ is a common element of the dependency sets of all the request types of the Example System. However, if Service $B$ is at fault, only requests of type $R_1$ will be impacted since Service $B$ belongs only to the dependency set of $R_1$.

The differential impact enables us to localize faults by observing the pattern of failures of request types. Figure 10 depicts the differential symptoms (impacts) of Figure 9 in an easy-to-understand tabular fashion. However, not all faults may produce distinctive patterns; the patterns depend on the system topology. For example, a failure in $N_1$ is indistinguishable from a failure in Service $A$ as they both produce the same observed pattern; both $N_1$ and $A$ are common elements of the dependency sets of the request types.

So far, we have considered only one fault at a time. However, in practice, multiple faults can co-occur. Though our methodology and its actual implementation in Instana easily generalizes to multiple independent concurrent faults, for the sake of simplicity, we focus essentially on one fault at a time in this paper.

| Y (observed measurements from availability dashboard) Request Success/Not anomalous (T=1\|F=0) | A | B | C | $N_1$ | $N_2$ |
|---|---|---|---|---|---|
| $Y_{R1}=0, Y_{R2}=0, Y_{R3}=0$ | 0 | 1 | 1 | 1 | 1 |
| $Y_{R1}=0, Y_{R2}=1, Y_{R3}=1$ | 1 | 0 | 1 | 1 | 1 |
| $Y_{R1}=1, Y_{R2}=0, Y_{R3}=0$ | 1 | 1 | 0 | 1 | 1 |
| $Y_{R1}=0, Y_{R2}=0, Y_{R3}=0$ | 1 | 1 | 1 | 0 | 1 |
| $Y_{R1}=0, Y_{R2}=1, Y_{R3}=0$ | 1 | 1 | 1 | 1 | 0 |

(Can't differentiate: rows 1 and 4)

**Figure 10. Symptom patterns corresponding to atomic faults**

The above scenarios provide an intuitive understanding of fault localization in complex systems. We've seen how the principle of spatial differential observability allows us to distinguish between different types of faults based on their impact patterns. We've also explored the complexities introduced by fault propagation and the importance of understanding system dependencies in the fault localization process.

## Fault localization using Boolean Satisfiability (SAT)

In our Example System, where we only allow fail-stop faults and work with the groups of assumptions A1 and A2, fault localization can be relatively straightforward. We can use the truth table of Figure 10. As mentioned earlier, in the table of Figure 10

- 1 represents the working state of a component,
- 0 represents a component's failed state and
- $Y_{R_j}$ represents observed measurements for the request type $R_j$.

The measurements mentioned above are obtained from tracing application requests. $Y_{R_j}$ can take the values of 1, representing success, or 0, indicating failure of the associated request type $R_j$.

As we can see, each unique combination of component states results in a distinct set of observations. In this idealized scenario, we can precisely determine which component has failed in some instances based on the observed measurements. For example:

- If $Y_{R_1} = 0, Y_{R_2} = 1$, and $Y_{R_3} = 0$, we can conclude that the component $N_2$ has failed.
- If $Y_{R_1} = 0, Y_{R_2} = 1$, and $Y_{R_3} = 1$, we can conclude that the component $B$ has failed.

As indicated earlier, the truth table in Figure 10 illustrates the principle of differential observability. Failure in $B$ is distinguishable from Failure in $C$ while failure in $A$ is indiguishable from failure in $N_1$. Precise fault localization is only possible when *different failure modes produce distinct observable patterns.*

In our deterministic system with only fail-stop failures, the fault localization problem can be viewed as a Boolean assignment problem, which can be formalized as follows.

1. Variables:
   - $X_i \in \{0, 1\} \; for \; i \in \{1, \ldots, N\}$, where $X_i = 0$ if the component $C_i$ is at fault (failed), and 1 if the component $C_i$ is working.
   - $Y_{R_j} \in \{0, 1\} \; for \; j \in \{1, \ldots, k\}$, $Y_{R_j} = 0$ if the request type $R_j$ fails, and 1 if the request type $R_j$ succeeds.

2. Constraints:
   For each request type $R_j$, the following Boolean assignment must be satisfied:

$$Y_{R_j} \Leftrightarrow \wedge_{i \in DependsOn \, (R_j)} \; X_i$$

3. Objective:
   Minimize the number of faulty components so that the Boolean constraints are satisfied. This translates to maximizing the number of working components under the constraints.

   maximize $\Sigma_{i=1}^{N} X_i$

### Insufficient spatial differentiability

Another limitation of the simplistic model is that there may not be sufficient spatial differentiability between fault scenarios. As shown in Figure 10, some failure patterns might be indistinguishable from others, making it challenging to pinpoint the exact source of the fault.

For example, if we observe $Y_{R_1} = 0$, $Y_{R_2} = 0$, and $Y_{R_3} = 0$, this could indicate a failure in the component $A$. It could also be a failure of the host $N_1$, or caused by a simultaneous failure of components $B$ and $C$. This lack of spatial differentiability is a common challenge in complex systems, and it often requires additional information or more sophisticated analysis techniques to locate faults.

The central question in the above situation is how to identify which sets of components

$$\{N_1\}, \{A\}, \{B, C\}, \{C, N_2\}, \{A, B, C\}, \{A, B, C, N_1\}, \text{or } \{A, B, C, N_1, N_2\}$$

are faulty. This is where our principle of differential observability is helpful. Let us assume for a moment that $A$ is the faulty component. In this case, collecting additional data can identify the real fault. For example, an SRE can provide feedback by adding additional measurements expressed in a functional notation in the following.

$$Y_{R_7} = f(N_1)$$

$Y_{R_8} = f(A)$,

where $\{N_1\}$ and $\{A\}$ are the singleton dependency sets corresponding to the request types $R_7$ and $R_8$. $R_7$ can be ping requests to the host $N_1$ and $R_8$ can be standard health checks on service $A$.

Measurements corresponding to the requests of types $R_7$ and $R_8$ will provide sufficient differentiability to identify the fault. Given that the fault is in $A$, measurements should indicate $Y_{R_7} = 1$ and $Y_{R_8} = 0$. This immediately helps to conclude that $N_1$ is working and $A$ has failed.

It is still unclear whether the components of the sets $\{B, C\}$ or $\{N_2, C\}$ failed simultaneously. Keeping with the principle of fixing one fault at a time, SREs will fix the fault in A and observe that none of $B$, $C$, or $N_2$ is in the failed state since the requests of all three categories $R_1$, $R_2$, and $R_3$, are succeeding. Such feedback can be collected automatically or manually by SREs. These fault scenarios illustrate the power of differential observability. In addition to localizing the fault, it helps identify the gaps in the observed measurements (symptoms). Additional data collection capability is available in Instana in the form of human feedback (thumbs up or down). It is for the SREs to collect the measurements and give our algorithm the required feedback to help further identify the faulty component.

The above discussion indicates the challenges of localizing faults in situations involving possible concurrent faults, which often occur in real life. Spatial differential observability combined with a) "Occam's razor for failures" and b) "Preferential failure bias" principles can significantly boost fault localization in such cases.

*Occam's razor for failures*

Occam's razor is the classical and well-honored problem-solving principle recommending searching for explanations constructed with the smallest possible set of elements. In case of failure analysis, this principle can be stated as follows.

Use the minimal set of atomic components to explain the observed symptoms as the probability of simultaneous multiple failures decreases rapidly as the cardinality of the set increases.

In our example system, the probability of two independent failures $\{B, C\}$ or $\{C, N_2\}$ occurring simultaneously is much lower than the probability of a single failure $A$ or $N_1$. According to Occam's razor, failure of $N_1$ or $A$ is more likely than the simultaneous failures of $B$ and $C$ (or $C$ and $N_2$). The failure of $A$ in isolation explains the observed symptoms $Y_{R_7} = 1$ and $Y_{R_8} = 0$. The fault localization using Boolean satisfiability described earlier, which *minimizes* the number of faulty components, adheres to Occam's razor for failures.

*Preferential failure bias (prior bias for failures)*

Existing publications, vendor specifications, environment-specific knowledge, or past data may provide the failure probabilities of the system components, which can be used as a prior bias for what is most likely to fail. For example, the probability of a host failure on a public cloud is much less than the failure of a service $A$ whose code has recently been updated. Hence, the failure diagnosis methodology should prefer $A$ over $N_1$ in such an environment.

Note that the above two principles are statistical, and one cannot completely rule out other failure modes. Instead, one can use these principles to rank the preference of best guess of fault given the symptom. For the scenario under discussion, the ranking of the probabilities of the component failures, and hence the ordering of investigations, can be expressed as $A > N_1 > \{B \wedge C\} \approx \{C \wedge N_2\}$. Such preferential bias cannot be modeled using the Boolean assignment formalism and requires probabilistic treatment, as detailed in a later section.

# Limitations of the simplistic model for fault localization

The simplistic and deterministic model fails to capture real-world behavior for several reasons:
1. **High request volume**: The Boolean SAT approach assumes only a single request per request type, which is unrealistic in most real-world systems. Consequently, we need a methodology that can model multiple requests for each request type and factor in all request types during failure diagnosis. In high-volume production environments, non-critical requests (e.g., logins and catalog browsing) typically outnumber critical ones (e.g., financial transactions and purchase orders). This disproportionate volume of non-critical traffic can obscure the impact of critical request failures by introducing noise and masking intermittent failures in critical operations.
2. **Non-fail-stop failures**: Not all failures result in a service's complete stoppage. Some may cause degraded performance or intermittent errors. For example, a memory leak or high

CPU spikes might cause sporadic failures that are difficult to correlate with a specific component.
3.  **Partial failures and environmental factors**: During a given interval, only a small percentage of requests of the same type may fail, resulting in nondeterministic behavior. This often occurs due to external factors like network latency, hardware problems, or general contention for computing resources. This partial failure mode is prevalent in distributed systems and can significantly complicate fault localization. The simplistic model does not consider partial system failures and nondeterministic behavior.
4.  **System complexity**: Real-world systems are often more complex and asynchronous than the atomic components shown in our example. They may include load balancers, caches, and other middleware that can obscure the true source of a fault.
5.  **State-dependent behavior**: Components can behave differently depending on their internal state or that of other components, resulting in context-dependent failure modes that aren't shown in a simple truth table. For instance, a database server might reject new requests if its request pools are full or if a firewall policy blocks traffic from certain clients.

In the following, we will build on this intuitive foundation to develop more formal methods for fault localization in complex systems. These methods will leverage the principles of differential observability and take into account the complexities of fault propagation that we have observed in the Example System depicted in Figure 6.

## Structural causal modeling

### Endogenous and exogenous variables

A **Structural Causal Model** (**SCM**) is a **Directed Acyclic Graph** (**DAG**) in which each variable is classified as **endogenous** (explained by the model) or **exogenous** (originating outside the model). The variables are nodes in the SCM DAG that connects the parent nodes to the child nodes, indicating probable causation.

- **Endogenous variables** (the ones we want to explain or predict) are assigned structural equations that depend on their parents.

- **Exogenous variables** capture external 'shocks' or influences not further decomposed in the model (e.g., random hardware faults, buggy updates, and ephemeral conditions).

Mathematically, each endogenous variable $V_i$ in the SCM has a structural equation of the form:
$V_i = f_i(Pa(V_i), U_i)$,
where:
- $Pa\ (V_i)$= the **parents** of $V_i$ in the DAG, i.e., the endogenous variables that directly affect $V_i$.
- $U_i$ = an **exogenous** variable (or 'shock') influencing $V_i$. Note that all exogenous factors are coalesced into one variable.

Specifically, if $V_B$ represents the health of service B, we may write $X_B := V_B$ to highlight that it is a binary state: either up (healthy) or down (failed) state. Recall that $X_B \in \{0,1\}$ is a random variable representing healthy (1) or failed (0) states. $X$ is a random variable because its value depends on random events. Consequently, the health of service $B$ can be modeled as: $X_B = f_B(Pa(X_B), U_B)$, indicating that $B$ could fail due to influences from components in $Pa(X_B)$, internal code bugs, or random external events. In the Boolean SAT formulation, $X_i$ was a pure Boolean variable. In the SCM approach, it continues to function as a 1 or 0 indicator for health versus failure, but later in the 'Probabilistic Model' section, we will associate a probability $\theta_i$ with it, where $\theta_i = P(X_i = 1)$ to capture the stochastic nature of $X_i$. Thus, the logical constraints from SAT become probabilistic constraints in the SCM.

Throughout this paper, we will often write $V_i$ for a generic endogenous variable in the DAG. When focusing on a concrete component like service B or host $N_1$, we will write $X_B$, $X_{N_1}$, etc. to emphasize the binary healthy or failed states. Both notations follow the same SCM definition; the difference is whether we are being generic or naming a specific component.

## Why SCMs for fault localization?

In distributed systems, *causation* is vital: we want to know whether a *specific* node or service's failure *causes* requests to fail. **Pearl's *do*-calculus** tells us how to evaluate "$do(X = 0)$" (i.e., forcibly setting the component's state X to 0, regardless of its usual probability) to see the *causal* impact on request outcomes. However, a *complete* SCM might be large and complex, especially if we have multiple layers (e.g., nodes, pods, services).

### *Example: Node and pod with partial dependency*

Consider a pod, $P_1$, that can only be scheduled on a Kubernetes host $N_1$. We can model the health of these entities as
1. Kubernetes host health: $X_{N_1} \in \{0, 1\}$
2. Pod health: $X_{P_1} \in \{0,1\}$

Each of the above two entities has an exogenous variable:
- $X_{N_1}$ depends on its exogenous shock $U_{N_1}$
- $X_{P_1}$ depends on $X_{N_1}$ and its exogenous shock $U_{P_1}$

Formally, in SCM notation:
- $X_{N_1} = f_{N_1}(U_{N_1})$, indicating the host $N_1$ is up or down purely due to external influences (e.g., random hardware fault, Operating Systems issues, etc.)
- $X_{P_1} = f_{P_1}(X_{N_1}, U_{P_1}) \Rightarrow X_{P_1} = 0$ if $X_{N_1} = 0$, otherwise $X_{P_1} = g_{P_1}(U_{P_1})$, indicating that the "pod is healthy (i.e., up)" ($X_{P_1} = 1$) requires both "host $N_1$ to be up" and "pod $P_1$'s exogenous shock not forcing $P_1$ down."

Feb 17, 2025                                                                                                                23

We can marginalize out exogenous factors as we are only interested in knowing whether a component is healthy or in a failed state by integrating over all possible exogenous variable $U_i$ weighted by the probability density function $p(U_i = u_i)$.

For example, we can define the probability of a Kubernetes host, $N_1$ being up as,
$$\theta_{N_1} = P(X_{N_1} = 1)$$
$$= \int 1[f_{N_1}(u_{N_1}) = 1] \, p(U_{N_1} = u_{N_1}) \, du_{N_1}$$

This integral (or sum) "marginalizes out" the exogenous factor $U_{N_1}$, leaving a single scalar probability $\theta_{N_1}$. The term $1[f_{N_1}(u_{N_1}) = 1]$ is an indicator function that is equal to 1 if $f_{N_1}(u_{N_1}) = 1$ (meaning $N_1$ is 'up') and 0 otherwise. So effectively, the above probability expression picks out the subset of exogenous states $u_{N_1}$ in which the host $N_1$ is healthy and then integrates (or sums up) their probabilities.

Similarly, we can define the probability of pod $P_1$ being up as,
$$P(X_{P_1} = 1) = P\left(X_{N_1} = 1 \wedge g_{P_1}(U_{P_1}) = 1\right)$$

Under typical independence assumptions (i.e., $U_{P_1}$ independent of $U_{N_1}$), we get
$$P\left(X_{P_1} = 1\right) = P\left(X_{N_1} = 1\right) \times P(g_{P_1}(U_{P_1}) = 1)$$
$= \theta_{N_1} \times \theta_{P_1|N_1=1}$, where
$$\theta_{P_1|N_1=1} = \int 1\left[g_{P_1}(u_{P_1}) = 1\right] p(U_{P_1} = u_{P_1}) du_{P_1}$$

However, when scheduling taints are not present and assuming there are no resource constraints or restrictions, it can be shown that $\theta_{N_1} = 1$, which leads to $P\left(X_{P_1} = 1\right) = \theta_{P_1|N_1=1}$. Note the use of Bayesian conditional probability in the above equation.

### Practical challenges to building a complete SCM

1. **Data collection overhead**: Including continuous metrics (CPU, memory, network latencies) for each host or container requires extensive instrumentation. Storing and analyzing large volumes of time-series data can be expensive.
2. **Complex functional forms**: Even if we collect CPU usage, the relationship between CPU usage and a service's failure probability can be nonlinear or domain-specific. Building robust structural equations or learning them in a data-driven manner is nontrivial.

As a result, for practical reasons, we choose to simplify. Instead of capturing partial or continuous failures, we treat each component as a binary "healthy or fail-stop" and ignore resource usage in the SCM DAG. Such simplification makes the causal modeling simpler; the inference algorithm tractable and sufficiently quick for real-time inference. It should be mentioned that Instana collects several important metrics and logs, which it independently analyzes to create events.

## SCM for the Example System

For our Example System of Figures 6 and 7, recall that each request type $R$ in our system depends on a specific set of atomic components (services or hosts) to succeed. As described earlier, we have:

- Request $R_1$: Replica of $A$ on $N_1$ calls $B$ on $N_2$.
    - Depends on $\{A, B, N_1, N_2\}$.
- Request $R_2$: Replica of $A$ on $N_1$ calls $C$ on $N_1$.
    - Depends on $\{A, C, N_1\}$.
- Request $R_3$: Replica of $A$ on $N_2$ calls $C$ on $N_1$.
    - Depends on $\{A, C, N_1, N_2\}$.

In a fail-stop setting, we can represent the state of a component $C_i$ ($i > 0$) with $X_i \in \{0, 1\}$, where $X_i$ is a binary random variable indicating whether the component $C_i$ is in healthy (1) or failed (0) states. The outcome 'success' or 'failure' of a request type $R_j$ is denoted by $Y_{R_j}$. Figure 8 visually illustrates the influence of each component $C_i$ on $Y_{R_j}$. We define this dependency as a 'DependsOn' relation and assert that any topological relation in a system can be converted to a set of DependsOn relations.

Representing the system in a set-theoretic manner for request types has several advantages:
- We should only model the dependency relations. In the real world, many relationships, such as 'runs on,' 'contains,' 'is a,' 'invokes,' etc., exist that are difficult to model in a uniform and simple way. For example, the [IBM Instana Observability platform](#) has numerous relations, and attempting to model each of them in terms of fault propagation is tedious and mathematically challenging. All of these relations can be converted to a straightforward dependency relationship when we restrict our focus to servicing various request types.
- It allows a unified representation of both service and infrastructure components.
- It is easily obtained using the run-time trace data provided by almost all monitoring tools used in practice.

A request fails if the state of *any* components in its dependency set is 0. Specifically, for a request type $R_j$, the outcome $Y_{R_j}$ is given by:

$$Y_{R_j} = \begin{cases} 1, & \text{if the state of all components in DependsOn}(R_j) \text{ are 1,} \\ 0, & \text{if at the state of at least one component in DependsOn}(R_j) \text{ is 0} \end{cases}$$

Equivalently, in algebraic form:

$$Y_{R_j} = \prod_{X_i \in DependsOn\ (R_j)} X_i$$

where each $X_i \in \{0, 1\}$.

If we think of each component $X_i$ as an endogenous variable with binary states, and each request outcome $Y_{R_j}$ as another endogenous variable, then the structural equation for $Y_{R_j}$ in the SCM formalism is:

$$Y_{R_j} = \prod_{X_i \in DependsOn(R_j)} X_i$$

Any exogenous factors that might cause components (e.g., $A, B, C, N_1, N_2$) of our Example System of Figure 6 to fail—such as random hardware errors on $N_2$, code bugs in B, etc. are *implicitly* wrapped into the probability $\theta_i = P(X_i = 1)$. By marginalizing out these external causes, we treat each $X_i$ as simply "healthy with probability $\theta_i = P(X_i = 1)$." Thus, the structural equation is $Y_{R_j} = \prod X_i$. Meanwhile $P\left(Y_{R_j} = 1\right) = \prod \theta_i$, where $\theta_i = P(X_i = 1)$. We will discuss how this can be used to infer the 'healthy probability' later in this paper.

## Interpreting causal insights

Although we no longer explicitly write out a full Structural Causal Model for each service or host, we retain the causal reasoning. For example, in our Example System illustrated in Figures 6 and 7, a low value of $\theta_B$, suggests that forcibly failing B ("$do(B = 0)$") *matches* the observed pattern of failing request $R_1$. If $R_2$ and $R_3$ succeed, it further confirms that a fault in $B$ (only relevant to $R_1$) is consistent with the data. On the other hand, if $R_1, R_2,$ and $R_3$ all fail, we might suspect that $\theta_A$ or $\theta_{N_1}$ are close to 0, since they appear in all dependency sets.

## The *do*-Operator: interventions vs. observations

**Pearl's *do*-calculus** formalizes the difference between:
- **Observing** $(X = x)$: "We notice that Service $B$ is failing, but it could be due to high CPU usage."
- **Forcing** $do(X = x)$: In practice, we cannot simply force real services to fail in production to gather data. Instead, we rely on the causal model to generate *synthetic* observations under *hypothetical interventions*. For example, we ask: "If we forcibly fail B (i.e., $do(B = 0)$) in our simulated SCM, which requests would fail?" If that simulated pattern of failures aligns with our observations, our posterior belief that B is faulty increases; if it does not match, we adjust that belief accordingly.

The ***do*-operator** 'breaks' the edges from X's causal parents, effectively overriding the usual dependencies. This concept is central to causal inference: "Would forcibly failing Service $B$ create the same request-failure pattern we observe?" If so, we conclude B's failure caused those request failures. If not, we rule out $B$. Such "what-if" scenarios are essential for systematically localizing faults.

In our causal inference framework, we use a **Beta-Binomial model** to iteratively generate these synthetic outcomes for $do(X = 0)$ or $do(X = 1)$ interventions—i.e., simulating the forcible setting of certain components to fail or succeed. Specifically, each component $C_i$ has a state $X_i$ drawn from a Beta-distributed prior $\theta_i$. Under an intervention, we fix $X_i = 0$ or $X_i = 1$ in the simulation and then predict how requests will succeed or fail.

Next, we compare the resulting synthetic request outcomes against real-time observations. This comparison can be formalized as minimizing a distance measure $D(observed\ data, synthetic\ data)$ over the space of possible interventions.

For instance, in a Bayesian setting, one might define the distance measure via the negative log-likelihood of the observed data given the current parameter estimates or via the Kullback–Leibler divergence between observed and generated outcome distributions. The inference process converges toward assigning $X_i = 1$ or $X_i = 0$, which most closely explains the observed request-failure patterns by iteratively updating each component's posterior distribution (i.e., recalculating the Beta parameters) to reduce this discrepancy. The convergence of the probabilistic inference approach is based on an earlier work on [failure localizations in communication networks](#).

In effect, we use **Pearl's *do*-operator** at each iteration to account for the causal influence of certain components (i.e., address confounders), generate hypothetical (synthetic) outcomes, and evaluate how well these hypothetical outcomes match the real failures and successes we observe. Through multiple updates, the algorithm identifies the components whose forced failures (or forced successes) best align with the observed error rates across the system. This approach enables us to systematically localize faults and refine our causal hypotheses in real-time.

## From full SCM to Beta-Binomial: a simplified model

### Probabilistic model

We adopt a probabilistic approach to overcome the limitations of the deterministic models described earlier. We use Beta distributions to model the uncertainty in each component's working condition. This approach enables us to generate and rank failure hypotheses based on observed request patterns and component dependencies.

### Beta distribution for component health

We begin by modeling the health state $X_i$ of each component $C_i$ with a probability $\theta_i$, which itself follows a Beta distribution

$$\theta_i \sim Beta(\alpha_i, \beta_i)$$

Here, $\theta_i$ encodes the probability that $C_i$ is functioning correctly and allows us to incorporate any "preferential failure bias" derived from empirical data or SRE knowledge. Conditioned on $\theta_i$, the observed binary health state $X_i$ follows a Bernoulli distribution:

$$X_i | \theta_i = Bernoulli\ (\theta_i)$$

In expectation terms, $\mathbb{E}[X_i | \theta_i] = \theta_i$, and the prior mean of $\theta_i$ is $\mathbb{E}[\theta_i] = \frac{\alpha_i}{\alpha_i + \beta_i}$ represents our prior estimate of $P(X_i = 1)$ and prior variance of $\theta_i$ is $var[\theta_i] = \frac{\alpha\beta}{(\alpha+\beta)^2\ (\alpha+\beta+1)}$. As new data (e.g., successes/failures) arrives, we update the parameters $\alpha_i$ and $\beta_i$ to form the posterior distribution, refining our estimate of each component's true health state.

The Beta distribution is particularly suitable for this purpose because:
1. It is defined on the interval [0, 1], matching our probability space for component health.
2. It can represent a wide range of shapes, allowing us to model various levels of certainty about a component's state.
3. It has convenient mathematical properties for Bayesian updating as we gather more information.
4. It is the conjugate prior for the Binomial distribution, which simplifies our inference process.

## Modeling request dependencies

We leverage the Structural Causal Model (SCM) for the request dependencies described earlier, but we adapt it for probabilistic inference. In this formulation, the state of each component $X_i$ is an endogenous variable with a binary state (healthy or failed). Its health probability, denoted by $\theta_i = P(X_i = 1)$, while any exogenous variables $U_i$ are integrated out (marginalized) to simplify the causal model. Consequently, the probability of a request $R_j$ succeeding is:

$$\gamma_j = P\left(Y_{R_j} = 1\right) = \prod_{\theta_i \in DependsOn(R_j)} \theta_i$$

which encodes an AND constraint among the components in the dependency set: **all** components must be healthy for the request to succeed. Recall from earlier discussions that the probability that a component $C_i$ is healthy is $\theta_i = P(X_i = 1)$.

In practical systems, however, other constraint types often arise:

1. **k-out-of-N constraint**: Only k out of N components must function. A common example is Apache Zookeeper, where the service remains operational if $k = \lfloor N/2 \rfloor + 1$ nodes are healthy.

$$P(Y_R = 1) = \sum_{i=k}^{N} \sum_{S \subseteq \{1,\ldots,N\} | |S|=i} \left[ \left(\prod_{m \in S} \theta_m\right) \left(\prod_{m \notin S} (1 - \theta_m)\right) \right]$$

For each subset S of dependencies (of size $\geq k$), the above expression represents the probability that subset S succeeds (first product term $\prod_{j \in S} \theta_j$) while the remaining dependencies fail (second product term $\prod_{j \notin S}(1 - \theta_j)$).

2. **OR constraint**: At least one out of N components must be healthy. This commonly appears in load balancing across multiple replicas of a stateless service.

$$P\left(Y_{R_j} = 1\right) = 1 - \prod_{i \in DependsOn\,(R_j)} (1 - \theta_i)$$

The necessity of these constraints depends on the observability and details of the system's topology. For instance, if request traces clearly show which replica is used, an *OR* constraint may not be necessary because each request path is explicit. Conversely, if tracing is inadequate, modeling an *OR* constraint using known (or automatically discovered) topology might be required. Similarly, k-out-of-N modeling may be unnecessary if the internal redundancy mechanisms (e.g., how Apache Zookeeper achieves quorum) are hidden from external services, which only need to know whether Zookeeper is 'available' overall. The decision to model or not model k-out-of-N is contingent on the level of granularity required for fault localization.

This multiplicative model captures the cascading effect of component failures on request success. Any component with a low health probability significantly reduces the overall request success probability.

## Modeling observed request successes

In real-world scenarios, we often do not directly observe whether each component is healthy. Instead, we rely on indirect indicators such as request successes or failures. By collecting traces (or spans), we can track which requests fail and which succeed, thereby inferring information about component health.

To formally link our model to observed data, we represent the total number of successful requests for each request type $R_j$ with a Binomial distribution. Previously, we used a binary variable $Y_{R_j}^k \in \{0, 1\}$ to denote the success or failure of the k-th request of type $R_j$. We now define the aggregate random variable

$$\overline{Y_{R_j}} = \sum_{k=1}^{n} Y_{R_j}^k,$$

which counts how many of those $n_j$ requests succeed. Under the assumption that each attempt is an independent trial with the same success probability $\gamma_j$, we have

$$\overline{Y_{R_j}} \sim Binomial\,(n_j, \gamma_j),$$

where $n_j$ is the total number of requests of type $R_j$, and $\gamma_j \in [0,1]$ is the probability that any individual request of type $R_j$ succeeds. A Binomial distribution is suitable because each request can be treated as a Bernoulli trial, and we are interested in the total number of successes across $n_j$ trials.

## Beta-Binomial model

Combining the Beta distribution for component health and the Binomial distribution for observed successes, we arrive at the Beta-Binomial model:

$$\theta_i \sim Beta(\alpha_i, \beta_i)$$
$$\gamma_j = \prod_{i \in DependsOn(R_j)} \theta_i$$
$$\overline{Y_{R_j}} \sim Binomial\ (n_j, \gamma_j)$$

This model allows us to incorporate both of our prior beliefs naturally about
- a component's health through the Beta distribution and
- our observed data through the Binomial likelihood.

The above formula captures the essence of Bayesian updating in the context of our model.

### *Prior Distribution Beta $(\alpha_i, \beta_i)$:*

Definition: Before observing any data, our belief about the health probability $\theta_i$ of component $C_i$ is encoded in a Beta distribution characterized by parameters $\alpha_i$ and $\beta_i$.

Interpretation: $\alpha_i$ can be viewed as the number of prior successes (e.g., healthy states), while $\beta_i$ represents the number of prior failures (e.g., failed states). The Beta distribution is a flexible prior distribution for probabilities, capable of expressing various degrees of certainty based on the choice of $\alpha_i$ and $\beta_i$.

### *Likelihood ($\overline{Y_{R_j}} \sim Binomial\ (n_j, \gamma_j)$):*

Definition: The observed data $\overline{Y_{R_j}}$ represents the total number of successful requests of type $R_j$ out of $n_j$ attempts. The success probability $\gamma_j$ for a request of type $R_j$ is a function of the health probabilities of its dependent components, specifically: $\gamma_j = \prod_{i \in DependsOn\ (R_j)} \theta_i$.

Interpretation: This models the AND constraint, in which all dependent components must be healthy for the request to succeed.

### *Posterior inference*

Combining our prior beliefs (encoded in the Beta distributions) with observed data (modeled by the Binomial distribution) allows us to perform Bayesian inference to update our beliefs about component health. The Beta-Binomial conjugacy allows for efficient posterior distribution computations.

The posterior distribution for each component's health will also be a Beta distribution with updated parameters:

$$\theta_i \mid \overline{Y}_{R_j} \sim Beta(\alpha_i', \beta_i')$$

where $\alpha_i'$ and $\beta_i'$ are the posterior shape parameters, updated based on the observed successes and failures. The symbol "|" means "given" or "conditioned on." $\theta_i \mid \overline{Y}_{R_j}$ reads as "the probability $\theta_i$ for the $i$-th component, after seeing the observed results $\overline{Y}_{R_j}$.

After observing $\overline{Y}_{R_j}$, we update our belief about $\theta_i$ using Bayesian inference. The posterior distribution remains a Beta distribution due to the conjugate prior relationship with the Binomial likelihood, with the parameters updated as follows:

- $\alpha_i' = \alpha_i + \#successes_i$
- $\beta_i' = \beta_i + \#failures_i$

$\alpha_i'$ incorporates prior successes ($\alpha_i$) and the new successes observed in the data. $\beta_i'$ incorporates prior failures ($\beta_i$) and the new failures observed. The symbol '#' represents 'number of'.

In practice, an iterative or factorized approach can determine how many new 'successes' or 'failures' to attribute to a given component $C_i$. For instance, if a request of type $R_j$ fails, we might increment the failure count for all components in the $DependsOn(R_j)$ set. More sophisticated schemes can weight these increments or update them iteratively in a way that best explains the observed pattern of successes or failures. Details of this factorization (i.e., inference algorithm) are beyond the scope of this paper. However, it suffices to say that we use **do-calculus** for modeling causality and **Markov Blankets** for real-time scalability.

The Beta distribution is conjugate to the Binomial distribution, meaning the posterior distribution is in the same family as the prior distribution. This property simplifies the computation, allowing for closed-form updates. Given the conjugate relationship, updating the posterior with new data involves simple arithmetic operations on the parameters, facilitating real-time inference. Finally, as more data is observed, $\alpha_i'$ and $\beta_i'$ are continuously updated, refining our estimates of $\theta_i$ to reflect the latest evidence. This online updating allows our model to adapt quickly to changing system conditions and provides a real-time estimate of component health. We have developed our domain-informed inference algorithm, which is about 1000 times more efficient in terms of time to converge than a naïve implementation available in standard Python Bayesian inference libraries.

After updating our beliefs, we rank components based on their probability of failure as mentioned in the following.

1. Calculate the mean $\mu_i$ and variance of $\theta_i$, the probability of $X_i$ of a component $C_i$.

2. Rank components in ascending order of $\mu_i$ – a lower value of $\mu_i$ indicates a higher probability of failure. The above ranking prioritizes the components to investigate, focusing attention to the most likely sources of failure.
3. The variance in the estimate of $\theta_i$ provides the uncertainty in our measure of $\theta_i$, and therefore, the uncertainty in our measure of $P(X_i = 1)$.

## Fault propagation path and causal inference

When ranking components, the fault propagation path must be considered. Components connected to the observed incident in the fault propagation path should be given higher priority.

In practice, multiple faults may need fixing due to propagation. However, in product implementations, it's often more manageable to approach this iteratively

1. Identify the most likely faulty component using the probabilistic model and causal graph.
2. Attempt to fix or mitigate the issue with this component.
3. Re-evaluate the system state and update component rankings.
4. Repeat steps 1-3 until the system returns to a healthy state.

This iterative approach allows for a systematic resolution of complex, multi-fault scenarios while prioritizing efforts on the most critical issues. It also aligns well with the practical constraints of system management, where it's often preferable to make incremental changes and observe their effects rather than attempting to resolve multiple issues simultaneously.

## Fault localization and hypothesis generation

Using the Beta-Binomial model, we can generate fault hypotheses by identifying components with low posterior health probabilities. After incorporating observed data, components with significantly reduced health estimates are prime candidates for being the root cause of system failures.

The output of this model is a ranked list of components, ordered by their likelihood of being the source of observed failures. This list serves as a set of failure hypotheses for further investigation and troubleshooting.

This mathematical framework provides a robust foundation for generating and ranking failure hypotheses in complex distributed systems. Incorporating prior knowledge, observed data, and component dependencies, our framework provides a theoretically sound approach to identifying potential root causes of system failures.

The Beta-Binomial model allows us to:
1. Represent uncertainty in component health.
2. Incorporate prior knowledge about system components, i.e., modeling the prior belief. For example, in the public cloud domain, host failure is less likely than service failure.

3. Update our beliefs based on observed request successes and failures.
4. Generate ranked lists of potential failure sources.

This probabilistic approach overcomes the limitations of deterministic models, offering a more nuanced and adaptable method for hypothesis generation in complex distributed systems. As we collect more data, the model continually refines its estimates, resulting in increasingly accurate failure predictions and more efficient troubleshooting processes.

## Temporal connections and hidden Markov models

It's important to note that the states of components are also connected in time, leading to a more complex model. While we won't delve into the full details of this temporal model in this paper, it's crucial to understand that a component's health state at time t is influenced by its state at time t-1.

This temporal dependency can be modeled using a **Hidden Markov Model** (**HMM**) structure:

$$P(X_i(t) \mid X_i(t-1)) = Transition\ Matrix$$

The Transition Matrix encodes the probability of transitioning between healthy and faulty states over time. The complete model, combining the Beta-Binomial emission probabilities with the HMM structure, allows us to capture both the uncertainty in our observations and the temporal evolution of component health states. While the HMM-based approach is beyond the scope of this paper, we incorporate a simple Markov chain for each component's state over time in production.

## Empirical illustration

We provide a straightforward illustration of our methodology and formalism using our Example System, as depicted in Figures 6 and 7. Suppose the workload on the Example System yields the following observations.
- $R_1$: 300 total requests, 60 successes, and 240 failures.
- $R_2$: 200 total requests, 40 successes, and 160 failures.
- $R_3$: 500 total requests, 100 successes, and 400 failures.

Let each $\theta_i$ start with a $Beta(0.1, 0.1)$ prior. This incorporates the following facts
- A component might be either "likely fully healthy or completely failing" (bimodal intuition).
    - Note that the mean of the Beta distribution here is $\alpha_i / (\alpha_i + \beta_i) = 0.1 / (0.1 + 0.1) = 0.5$, implying equal probabilities of success or failure.
- The confidence in the mean is extremely low.
    - The variance of the Beta distribution here is $(\alpha_i * \beta_i) / ((\alpha_i + \beta_i) ** 2 + (\alpha_i + \beta_i + 1)) = (0.1 * 0.1) / ((0.1 + 0.1) ** 2 * (0.1 + 0.1 + 1)) = 0.01 / (0.04 * 1.2) = 0.01 /

.048 ≈ 0.2083, which is a very low value indicating extremely low confidence in the mean.

Using the inference algorithm described earlier, our causal model converges to
- $\mu_B \approx 0.90$,
- $\mu_A \approx 0.95$,
- $\mu_C \approx 0.95$,
- $\mu_{N_1} \approx 0.10$,
- $\mu_{N_2} \approx 0.90$

Here, $N_1$ is suspect ($\mu_{N_1} = 0.10$), signifying that $N_1$ is more likely failing in the observed time window than the other components. This outcome arises naturally because **every** request type under consideration ($R_1, R_2$ and $R_3$) depends on $N_1$, and each has a high failure rate. Under our causal model's logic:
- If $N_1$ is indeed failing (low $\mu_{N_1}$), *all* three request types fail with high probability, which *matches* the observed data.
- Meanwhile, $\mu_B$ and $\mu_C$ remain relatively high (~0.95) because there is *no* unique evidence that $B$ or $C$ fails beyond the widespread failures seen by all requests ($R_1, R_2$ and $R_3$).
- $N_2$ also stays at ~0.90 because having *all* requests fail that heavily is better explained by an issue on $N_1$ (common to $R_1$, $R_2$ and $R_3$) rather than $N_2$ alone (common to $R_1$ and $R_3$, but not $R_2$).

In our Example System, the Beta-Binomial mechanism illustrates how a *common fault* in one shared component ($N_1$) can be pinpointed by the consistently high failure rate of each request path that depends on it. Of course, in a real system, an SRE would confirm this suspicion with logs or further testing; however $\mu_{N_1} \approx 0.1$, the mean of the posterior probability provides a strong automated signal that "$N_1$ is likely the culprit."

## Limitations and future extensions

We made several simplifying assumptions to showcase the core idea. Most of these assumptions are relaxed in our actual implementation except the following two assumptions:

## Continuous metrics and confounders

Our approach ignores CPU usage, memory saturation, and network metrics. These factors can *cause* partial failures in real life, so ignoring them can lead to misleading attributions. In a complete SCM, we would add nodes for CPU usage, but we must define or learn how CPU usage influences $\theta_i$. This is more domain-specific and can be much more data-intensive. In reality, learning this influence function is not practical, and this influence function is not universally applicable across all applications. Consequently, we simplified our approach by disregarding resource consumption in general. Such simplification makes the causal modeling simpler, and the inference algorithm tractable and efficient for real-time inferences. This does not imply that Instana neglects other datasets. As previously stated, Instana collects several important metrics and logs. These metrics and logs are analyzed independently to create events using anomaly detection algorithms or from known problems (e.g., regex log patterns). These events surface as associated events for the identified faulty entity.

### Observational coverage

The Beta-Binomial model offers limited insight when specific requests are rarely observed or when partial coverage leaves some components untested. It is critical to ensure that the system's request traffic *sufficiently covers* all relevant paths. Otherwise, multiple faults remain indistinguishable (e.g., Host $N_1$ vs. Service A in our example) if no requests differentiate between them. We address these limitations by collecting real-time feedback from SREs to bias our models to the actual failing components.

# A summary of Instana's implementation of the causal AI-driven probable root cause identification

The following provides a simplified overview of Instana's root cause identification implementation. Note that we ignored many implementation details for ease of understanding, brevity, and simplicity.

Instana automatically instruments and collects *trace*s for all the supported technologies. A trace represents a single request and its journey through a system of services. It is a collection of operations that document the lifecycle of a request as it moves through a distributed system. Each trace consists of one or more *call*s that represent communications between two services – a request and a response. Instana captures the call data both on the caller and the callee side. The timings of code executions are called *span*s, which are actions with start and end times. Each operation within a trace is a span, that records details like the operation name, start and end times, and contextualized metadata such as the HTTP method or status code, all added to the spans by Instana agents. Like other distributed tracing systems, Instana's tracing is also based on Google's [Dapper](#). For further details, refer to the [Concepts of tracing](#) section of the Instana documentation.

Instana also automatically creates and dynamically updates topological graphs of applications' dependencies. These graphs encompass physical components like hosts, databases, etc., along with logical components like traces, services, etc. Instana dynamically updates the states of the components, such as metric and configuration data. We can logically think of two dynamic graphs: one related to infrastructure components – refer to the [Using the dynamic graph](#) of Instana documentation; and the other dealing with application-centric components – refer to Figure 11 of this paper. Our algorithm uses both these dynamic graphs for its analysis of fault localization.

The tracing and topology data collected by Instana serve as the foundation for implementing the causal-AI algorithm for probable root cause analysis. By capturing comprehensive trace information and maintaining a dynamic, real-time view of infrastructure and application relationships, Instana provides the necessary context for understanding the interdependencies and behaviors of system components.

To model the health state of every observed entity using the formalisms described earlier in this paper, let's first define the scope of consideration—what components can be identified as at fault? Generally, we consider infrastructure and application components as candidates in the probable root cause identification process.

We use the principle of spatial differential observability to localize faults. As detailed in this paper's "Insufficient spatial differentiability" section, localizing faults to atomic components is often impossible without additional measurements through special request types or SRE feedback. In those cases, our algorithm logically merges components to localize faults.

For example, consider a pod containing a single container running a microservice component of an application. Without other data helping spatial differentiability, our algorithm logically merges the pod, the container, and the corresponding process into one component for fault localization. If the process fails, the container and microservice are down, too.

In actual implementation, we collect and enrich calls at regular intervals with infrastructure topologies such as source and destination, endpoints, processes, etc. As discussed above, we also merge infrastructure components that are not differentially observable. For each request type $R_j$, we aggregate the total number of successful and failed requests corresponding to $R_j$ within the above-mentioned time interval. In Instana, request types correspond to the concept of 'calls,' which consists of requests and their corresponding responses.

$$\overline{Y_{R_j}} = count\ of\ succesfull\ calls = \sum_{k=1}^{N} Y_{R_j}^k = \left( \sum_{k=1}^{N} \prod_{\theta_i \in DependsOn(R_j)} \theta_i \right) = \left( \sum_{k=1}^{N} \prod_{\theta_i \in (\{caller\ components\} \cup \{callee\ components\})} \theta_i \right)$$

where, $\overline{Y_{R_j}}$ is the count of successful requests of type $R_j$ (i.e., calls), N = the total number of requests of type $R_j$, $DependsOn(R_j)$ includes both the caller and callee component sets consisting of differentially observable components on the source and destination sides corresponding to the request type $R_j$.

Based on the health information of the request types, for each component $C_i$ in the caller or called component sets of all the request types $R_j$s, our probabilistic fault localization algorithm computes $\mu_i$ the mean of the Beta-distribution modeling probability that the component $C_i$ under consideration is healthy. As mentioned earlier, the algorithm also ranks the components according to their associated $\mu$ values in an ascending order.

# Causal AI-driven probable root cause identification in Instana user interface

The AI-infused Root Cause Identification (RCI) feature is a powerful new addition to Instana, probably distinguishing it from other monitoring tools available in the market.

The internal fault diagnosis mechanism of the previous generation of Instana, and the AI-infused Instana's RCI analysis hinge on the dynamic directed call graph -- also known as 'Dependencies' in the Instana GUI or simply dynamic graph in Instana literature. A call graph of Instana visually displays
- all the individual components of the application, along with their names as the nodes of the graph,
- the connectivity between the individual components as edges and
- the direction of the caller-callee request flow as the direction of the edges of the graph

The graph is potentially dynamic since depending on the user's interaction with the system
- the graph and the connectivity of the nodes through edges may evolve gradually
    - new directed edges can appear,
    - new nodes in application components can also appear, along with new directed edges connecting them with other existing or just added nodes,
    - existing nodes and edges can disappear from the call graph if the corresponding paths remain dormant for a certain period.

Figure 11 is a simple call graph of an application that graphically displays the components of a web application. The directed nature of the graph does not show up in a static screen capture – Instana GUI provides animation to indicate the direction of the traffic flow between components, which is the directionality of the edges between the graph's connected nodes.

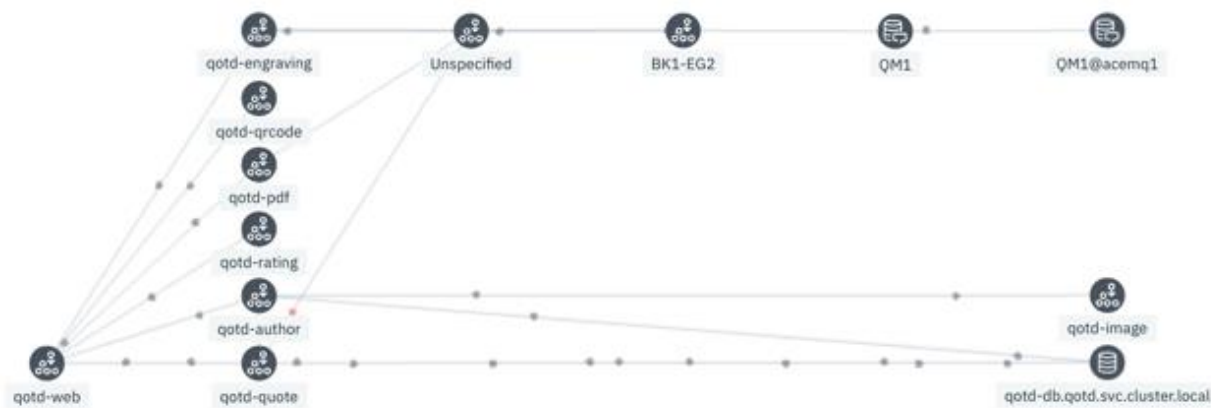

**Figure 11. A call graph of an application**

The non-AI-based root cause analysis of the previous generation of Instana can start from several places in the Instana GUI:
- an Instana-generated 'Alert,' it's 'Triggering event,' and the 'Related events' of the triggering event, or
- 'Issues' noticed in one or more components in the call graph or
- directly from the 'Events' emitted from the alert-generating component of a call graph.

Regardless of the starting point, to detect the root cause, SREs typically follow standard debugging techniques: analyze calls associated with events under consideration, carefully drill down the call stack, analyze traces, and relate calls to log files, error messages, and available stack traces. Although Instana GUI provides significant help to SREs in troubleshooting, the entire process is manual, requires expertise and experience from SREs, and can be tedious, time-consuming, and error-prone.

Figure 12 shows the events 'Related' to a 'Triggering' event corresponding to an Instana-generated 'Alert' in an application component.

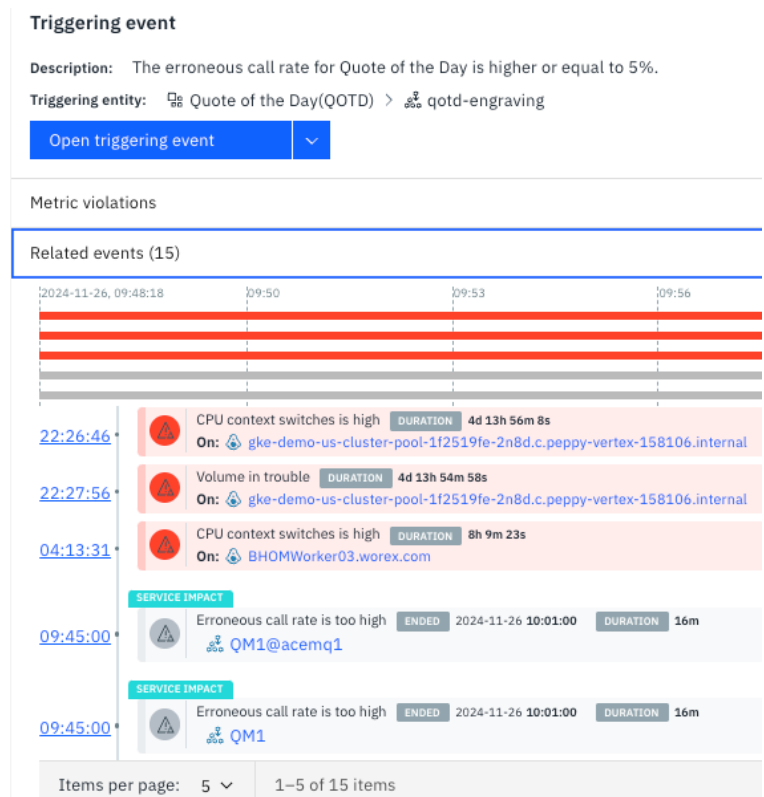

**Figure 12. Related events in Instana**

The relationship among the related events in Figure 12 is spatiotemporal. Instana essentially will correlate two events, say Event$_A$ and Event$_B$, if they occur in connected nodes in the call graph within a specific time range of several minutes. In the formal discipline of Statistics, 'correlation' does not imply 'causation,' which means in this context that there may not be any causal relationship between two events correlated by Instana – Event$_A$ may not be the cause of Event$_B$ or vice versa. The root cause analysis using related events may yield erroneous results without a causal relationship.

Unlike the manual, time-consuming endeavor, the causal AI-infused Instana automatically identifies the root cause entities with high precision and adequate explanation in near real-time using AI on internally constructed causal graphs and other theoretically sound and well-respected formalisms discussed in this paper. The root cause component can be associated with any node on the call graph of Figure 11: a process, an infrastructure entity, or an endpoint of an application's component in the call graph. The root cause identification (RCI) process has a low memory footprint and does not consume significant computing resources.

As mentioned earlier, for each element $C_i$ of the set of possible faulty components, our probabilistic RCI algorithm computes and associates $\mu_i$ with it. $\mu_i$ being the mean of the Beta-distribution modeling probability of $C_i$'s healthiness. Note that for a component $C_i$, a lower value of the corresponding $\mu_i$ indicates a higher probability of failure. For a set of probable root cause components, the RCI algorithm also sorts the conceptual tuples <$C_i$, $\mu_i$> in ascending order of $\mu_i$. A component $C_i$ is only considered for exposition in the Instana probable root cause GUI, if it's associated $\mu_i$ is less than an internal threshold value $\mu_T$. The top three components in the sorted list associated with the lowest $\mu$ values all less than $\mu_T$ are exposed as probable root causes in the Instana GUI.

The selected probable root causes, if any, are displayed in ascending order of their $\mu_i$ values from left to right; the root cause component $C_i$ associated with the lowest mean ($\mu_i$) value of the set of components under consideration occupies the leftmost position in the GUI. Instana divides the $\mu_i$ values associated with the possible faulty components that are less than the threshold value $\mu_T$ into three disjoint categories: High, Moderate, and Low, according to their failure chances -- ($\mu_{High} < \mu_{Moderate} < \mu_{Low}) < \mu_T$. For a triggering event, the RCI algorithm may identify more than one root cause whose associated $\mu$ value belongs to the same category.

Figure 13 is a typical screen capture displaying probable root causes in the Instana GUI. Though all three mean values associated with the top three probable root causes corresponding to the three tabs in Figure 13 belong to the Moderate category, based on their associated individual $\mu_i$ values, the probable root causes are arranged in ascending order from left to right; the leftmost one associated with the lowest mean is explicitly labeled as 'Most likely cause.'

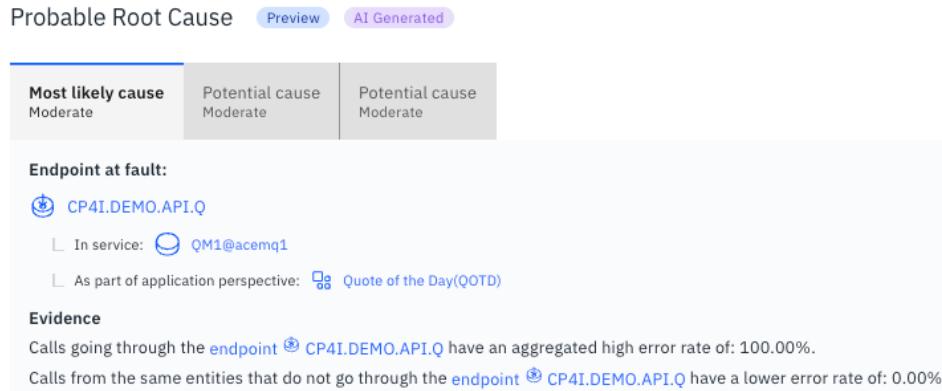

Figure 13. Probable root causes

The number of potential root causes identified by the RCI algorithm, which have associated mean values less than the internal threshold, may be fewer than three. Figure 14 shows a situation where only one probable root cause is displayed. In extreme cases, no probable root cause may be displayed, implying that the algorithm failed to identify any probable root cause with a reasonable degree of certainty. This may happen for reasons like lack of detailed traces. To minimize the chance of false positives, which may negatively affect the user's confidence, Instana does not surface any root cause if none of the associated mean values of the identified root causes are less than the internal threshold.

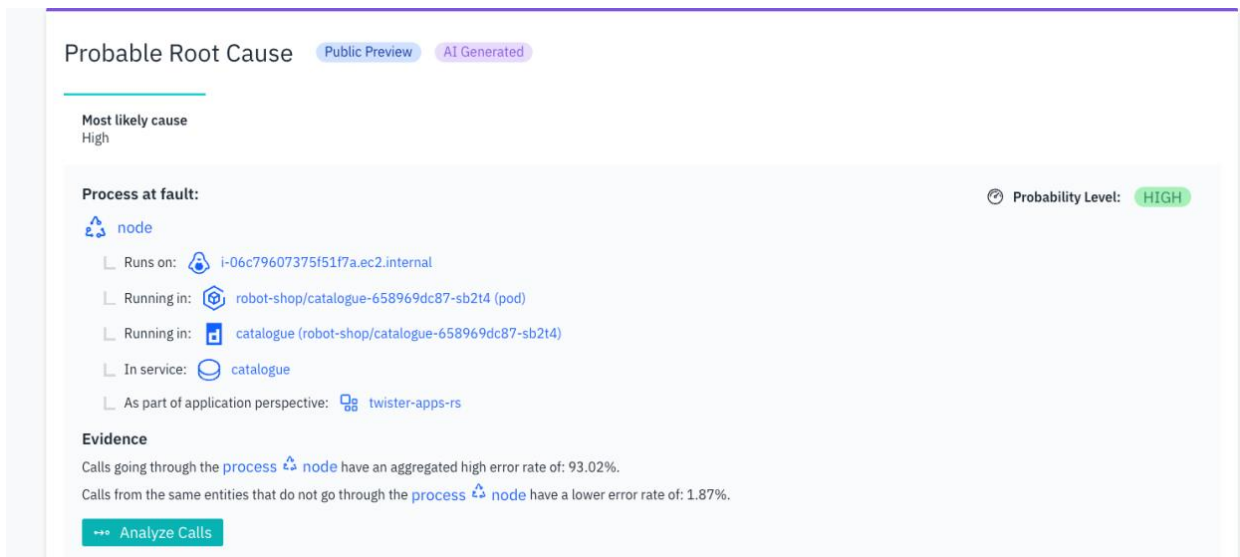

Figure 14. Only one probable root cause

The causal AI for RCI provides explanations in the 'Evidence' paragraph of the corresponding root case, as shown in Figures 13 and 14. Instana also offers a detailed spatial viewpoint of the root cause component, as depicted in Figures 13 and 14. In Figure 13, Instana gives context for the `CP4I.DEMO.API.Q` Endpoint root cause component by detailing that it is an Endpoint of Service `QM1@acemq1` of the `Quote of the Day (QTOD)` application. This context can be immensely helpful to SREs while troubleshooting and fixing faults in real life. For example, Figure 14 provides a detailed hierarchical spatial context for the root cause component—a node.js process. The node.js process in the `catalogue` service component of the `twister-app-rs` application executes in the `catalogue(robot-shop/catalogue-658969dc87-sb2t4)` container in the `robot-shop/catalogue-658969dc87-sb2t4` pod hosted in the `i-06c79---a.ec2.internal` cloud virtual machine (VM). The detailed hierarchy in Figure 14 also showcases the power of RCI in pinpointing the faulty infrastructural component. In the example of Figure 14, the `catalogue` service is realized by three containers in three pods residing in three worker nodes hosted in three cloud VMs. The simulated fault happened in one of the containers, and Instana's RCI algorithm accurately identified the fault-causing container, leaving the other two non-faulty ones. Note that Figure 14 displays the root cause in the Instana GUI related to the similar example discussed in Example 2 of the 'Motivating examples' section of this paper.

A detailed and precise root cause identification enables an SRE to quickly focus on the faulty component and take appropriate actions to resolve the issue. Corrective actions can vary based on the situation; they may involve a simple restart or a restart that requires reconfiguration or emergency code changes to the faulty component. The corrective action might necessitate a thorough analysis of diagnostic logs, events, and detailed traces related to the faulty component by the SRE before determining the necessary corrective measures. An SRE is also expected to consult and potentially utilize AI-generated remediation actions associated with the triggering event within Instana's remediation framework. For more information, refer to the [Intelligent Remediation](#) section of the Instana documentation.  Based on our study in real-life or simulated situations, we can safely conclude that the newly introduced root cause identification capability of the AI-infused Instana can reduce the Mean Time To Recovery (MTTR) by at least 80%.

Figures 13 and 14 indicate that Instana's AI-infused RCI algorithm is currently in the Public Preview phase. Users can provide feedback regarding its accuracy and helpfulness by clicking the thumbs-up or thumbs-down buttons in the Instana GUI. The Instana team uses the feedback to improve the algorithm. IBM intends to make the AI-infused RCI generally available in early 2025.

# Real-life experiences with Instana's casual AI-based root cause identification

From its initial design and implementation, our RCI algorithm continued to show great promise and excelled at assisting SREs in almost all internal and external environments. The following mentions two real-life situations with severe service outages in which the algorithm exemplifies its potential value in identifying faulty components quickly and accurately in large, complex environments.

When our RCI algorithm was still under development, we tried it internally, involving partial service failure in a very large distributed environment. Experienced SREs followed traditional troubleshooting techniques, using the monitoring and telemetry data of the system obtained from Prometheus and Grafana. It was a very long-drawn troubleshooting process that involved several erroneous root cause identifications in the course. It took more than twelve hours for the SREs to arrive at the right faulty component. On the side, after the environment was stabilized, we tried our RCI algorithm on the making using the collected traces, and it successfully pinpointed a software networking switch as the component at fault in less than five minutes. Indeed, the service was completely restored after a simple change to the configuration of the software switch.

A severe system slowdown problem was encountered in an internal [Sterling Order Management System](#) (OMS) environment. Experienced OMS SMEs using the traditional manual fault detection approach of analyzing single traces concluded after a significant amount of time that the DB2 database component was at fault because of its configured rate limit violations. In contrast, Instana's RCI implementation automatically detected the component at fault to be a WebSphere Application Server (WAS) instance connected to the DB2 database. Further analysis from logs and metric data confirmed our RCI's algorithm's correctness – indeed, the WAS instance was at fault, not the DB2 instance; the WAS instance exhausted its configured thread pool. We have already discussed this interesting real-life scenario in Example 1 of the 'Motivating examples' section of this paper.

We also verified and validated the effectiveness of our causal AI-based RCI algorithm and implementation in several customer scenarios, working with their SREs in external environments of varying complexity. In all cases, when sufficient telemetry data was available, our RCI implementation could correctly identify the possible components at fault. In instances where telemetry data was sparse, Instana's RCI implementation did not surface any faulty component because of a perceived lack of certainty, as indicated earlier.

# Putting Instana's RCI in perspective

Modern enterprise applications are becoming more distributed, scalable, fault-tolerant, highly available, and hybrid, spanning cloud and on-prem components. These applications are fast embracing and incorporating newer technologies like containerization, Kubernetes, AI, etc. The traditional monitoring tools that served their purpose well earlier have not kept up with the rapid technological advances of highly dynamic distributed hybrid applications.

Many vendors currently offer Application Performance Monitoring (APM) and observability utilities for enterprise applications. The commercially available APM and observability products can be roughly divided into five categories.
- Traditional APM tools collect system performance data and display them flexibly to users. They also produce reports to track performance and availability data trends. SREs use all this

- information for troubleshooting. However, as mentioned earlier, these tools may not generally be adequate for modern highly distributed hybrid applications.
- A class of APM tools performs anomaly detection on the collected performance data and traces. Events and alerts are generated using one or more anomaly detection techniques. However, these tools can generate significant noise by creating 'events storms,' which can seriously impede root cause detection in real-life troubleshooting.
- To reduce noise and help SREs focus on important events and alerts, another class of observability vendors uses correlation coefficients and machine learning models to correlate components for their root cause analyses. As we mentioned earlier in this paper, correlation is not causation. While some of these tools are popular in the field, their theoretical basis is weaker than that of Instana, which is based on causality. Our initial observations conform to the above statement.
- A fourth class of APM tools tries to learn models by identifying the root cause from known fingerprints of symptoms curated manually. This class of observability tools uses built-in models of dependent variables, which we informally call patterns, to accurately identify *common* root causes in cloud-native environments. This approach is interesting and may quickly determine the correct root causes in some cases. However, due to the fundamentally static nature of the built-in models of patterns, this approach may face significant issues in real-life scenarios.
    - After extensive efforts, it is possible to enumerate common patterns involving popular middleware and other systems hosting standard microservice applications. However, anticipating and preloading pattern configurations for every possible middleware, system, application, and runtime environment is impractical. All software, including middleware, databases, customer applications, and others, undergo change—bugs are fixed, enhancements are made, and new specifications are implemented. These modifications can alter existing symptoms, remove some of them, or introduce new ones along with their altered or new associated fingerprints. Considering the potential combinations of software components and the inherent dynamism of their life cycles, built-in pattern models may often require careful modifications and expansions to include the diverse components and contexts encountered in large-scale enterprise deployments. This process can be labor-intensive, costly, and prone to errors.
    - Even in relatively simple cases, the pre-populated built-in models of patterns may not fully account for the possible existence of malicious (erroneous) components that can act as confounders, resulting in the erroneous identification of root causes.
- A causal-AI based observability platform, like IBM Instana. Instana uses the Pearlean framework for its causal inference described in this paper. Given sufficient trace data, Instana's probabilistic algorithm will identify root causes with high precision in near real-time without consuming significant computing resources for all types of middleware, systems, and applications in any environment: cloud-native, traditional, or hybrid.

Because of its powerful features, including its root cause detection ability, Instana's selection as [CRN's 2024 Product Of The Year](#) in the Application Performance and Observability category is not surprising.

# Conclusions

The probabilistic approach to fault localization presented in this paper provides a powerful framework for dealing with the complexities and uncertainties inherent in modern distributed systems. By leveraging beta distributions, causal graphs, and online inference techniques, we can perform effective fault localization even in environments with partial observability and complex failure modes.

Future research in this field can be significantly enriched by harnessing both agent-based methodologies and generative AI. For instance, incorporating advanced machine learning techniques—particularly those involving generative models—holds promise for improving causal graph generation and parameter estimation by automatically identifying complex patterns in system behavior. In parallel, agent-based approaches can enable adaptive and collaborative fault localization strategies, especially when paired with reinforcement learning algorithms that evolve over time. Another promising direction lies in integrating natural language processing techniques within intelligent agents to seamlessly include unstructured data sources, such as logs and error messages, into the probabilistic model. By blending these innovative methods—machine learning, generative AI, agent-based systems, and NLP—the field stands to achieve more robust, efficient, and adaptive solutions for fault localization and system diagnostics

As systems grow in complexity, advanced fault localization techniques will become increasingly critical for maintaining the reliability and availability of digital services. The concepts and methods presented in this paper provide a foundation for addressing these challenges and pave the way for further innovations in dependable computing.

## Acknowledgments

The authors gratefully acknowledge the technical contributions, valuable discussion sessions, and all the support received from Marc Palaci-Olgun, Brad Blancett, Arthur de Magalhases of Instana development; Daby Sow, Ph. D., Director, Ruchir Puri, Ph. D., Chief Scientist, Fellow, and VP, Nicholas Fuller, Ph. D., VP of IBM Research; Danilo Florissi, Ph. D., VP of Instana and Turbonomic development.

Debasish acknowledges the support of Om Bachu, the founder of Guild Systems Inc. Debasish fondly remembers the support and encouragement of Laura Scott, the retired IBM CSM VP, and Aboud Ghazi, the IBM US Industry CSM leader.